\documentclass[%
 reprint,
 amsmath,amssymb,
 aps,
]{revtex4-2}

\usepackage{graphicx}% Include figure files
\usepackage{dcolumn}% Align table columns on decimal point
\usepackage{bm}% bold math
\usepackage{here}
\begin{document}

\preprint{APS/123-QED}
\title{Predicting unobserved climate time series data at distant areas via spatial correlation using reservoir computing}%

\author{Shihori Koyama}
\affiliation{Toyota Central R\&D Labs., Inc., Nagakute, Aichi 480-1192, Japan}
\email{shihori-koyama@mosk.tytlabs.co.jp}
\author{Daisuke Inoue}
\affiliation{Toyota Central R\&D Labs., Inc., Nagakute, Aichi 480-1192, Japan}
\author{Hiroaki Yoshida}
\affiliation{Toyota Central R\&D Labs., Inc., Nagakute, Aichi 480-1192, Japan}
\author{Kazuyuki Aihara}
\affiliation{International Research Center for Neurointelligence, The University of Tokyo, Bunkyo, Tokyo 113-0033, Japan}
\author{Gouhei Tanaka}
\affiliation{International Research Center for Neurointelligence, The University of Tokyo, Bunkyo, Tokyo 113-0033, Japan}
\affiliation{Graduate School of Engineering, Nagoya Institute of Technology, Nagoya, Aichi 466-8555, Japan}

%\tableofcontents
\begin{abstract}
Collecting time series data spatially distributed in many locations is often important for analyzing climate change and its impacts on ecosystems. However, comprehensive spatial data collection is not always feasible, requiring us to predict climate variables at some locations. 
This study focuses on a prediction of climatic elements, specifically near-surface temperature and pressure, at a target location apart from a data observation point. 
Our approach uses two prediction methods: reservoir computing (RC), known as a machine learning framework with low computational requirements, and vector autoregression models (VAR), recognized as a statistical method for analyzing time series data.
Our results show that the accuracy of the predictions degrades with the distance between the observation and target locations. We quantitatively estimate the distance in which effective predictions are possible. 
We also find that in the context of climate data, a geographical distance is associated with data correlation, and a strong data correlation significantly improves the prediction accuracy with RC. 
In particular, RC outperforms VAR in predicting highly correlated data within the predictive range.
These findings suggest that machine learning-based methods can be used more effectively to predict climatic elements in remote locations by assessing the distance to them from the data observation point in advance.
Our study on low-cost and accurate prediction of climate variables has significant value for climate change strategies.
\end{abstract}
%TC:endignore
\maketitle
\section{\label{sec:introduction}Introduction}
Climate change is making a major impact on the global environment and causing serious problems in ecosystems and human life. To clarify the causes and effects of climate change and take appropriate measures, information from actual observations of climatic elements plays an important role. Therefore, it is desired to acquire observation data at as many locations as possible in order to obtain more accurate information. In practice, however, it is often unfeasible to obtain data from spatially uniform observation points because of a difficulty in the acquisition of observation data and/or the economic cost of equipment and installation. Thus, it is important to estimate the values of climatic elements that cannot be observed with a high degree of accuracy.
One estimation method is a spatio-temporal interpolation of climate data based on data assimilation using numerical models and observational data~\cite{CBB+2018}. These approaches have been traditionally used in meteorology because of their superior accuracy, but their large computational requirements make them unsuitable for online interpolation.

As another approach, a growing number of machine learning methods have been developed in recent years, which learn from observable time-series data and subsequently predict unobserved climatic elements and their future values. For example, they have been used for the prediction of climatic variables such as temperature and pressure~\cite{ACR+2016,HQZ+2018,MOF2021,KMA+2021,CMH+2022,PSH+2022,XYH+2018,Nadiga2021,MOF2022,PW2022}, the detection of phenomena and patterns associated with extreme weather such as cyclones and atmospheric blocking phenomena~\cite{Stott2016,KCH+2020,SKG2020,TVL+2021,MBK+2021,SNY+2022}, and the applications related to climatic factors such as predictions of crop yield ~\cite{VMS2014,GFH+2021} and water temperature in lakes~\cite{WRA+2021}. These approaches are well suited to climate data, where immediacy of prediction is required, because once model parameters are learned from the data they can be used to make fast predictions. In practical applications of these approaches, it is desirable to fully interpolate missing or uncertain observation data with data at a minimum number of observation points. Once the optimal observation points to achieve this is specified, the number of excessive observation points can be reduced, thereby reducing installation costs, maintenance effort, data processing, and energy consumption. To design such optimal observation points, it is necessary to carefully examine the conditions that can guarantee an adequate accuracy of the prediction. Although predicting future states at data observation points are common in previous studies, there has been few detailed discussions on the extent to which spatially extended physical variables can be predicted from the observables at remote points.

\begin{figure}[htb]
  \centering
  \includegraphics[keepaspectratio, width=.95\columnwidth]
  {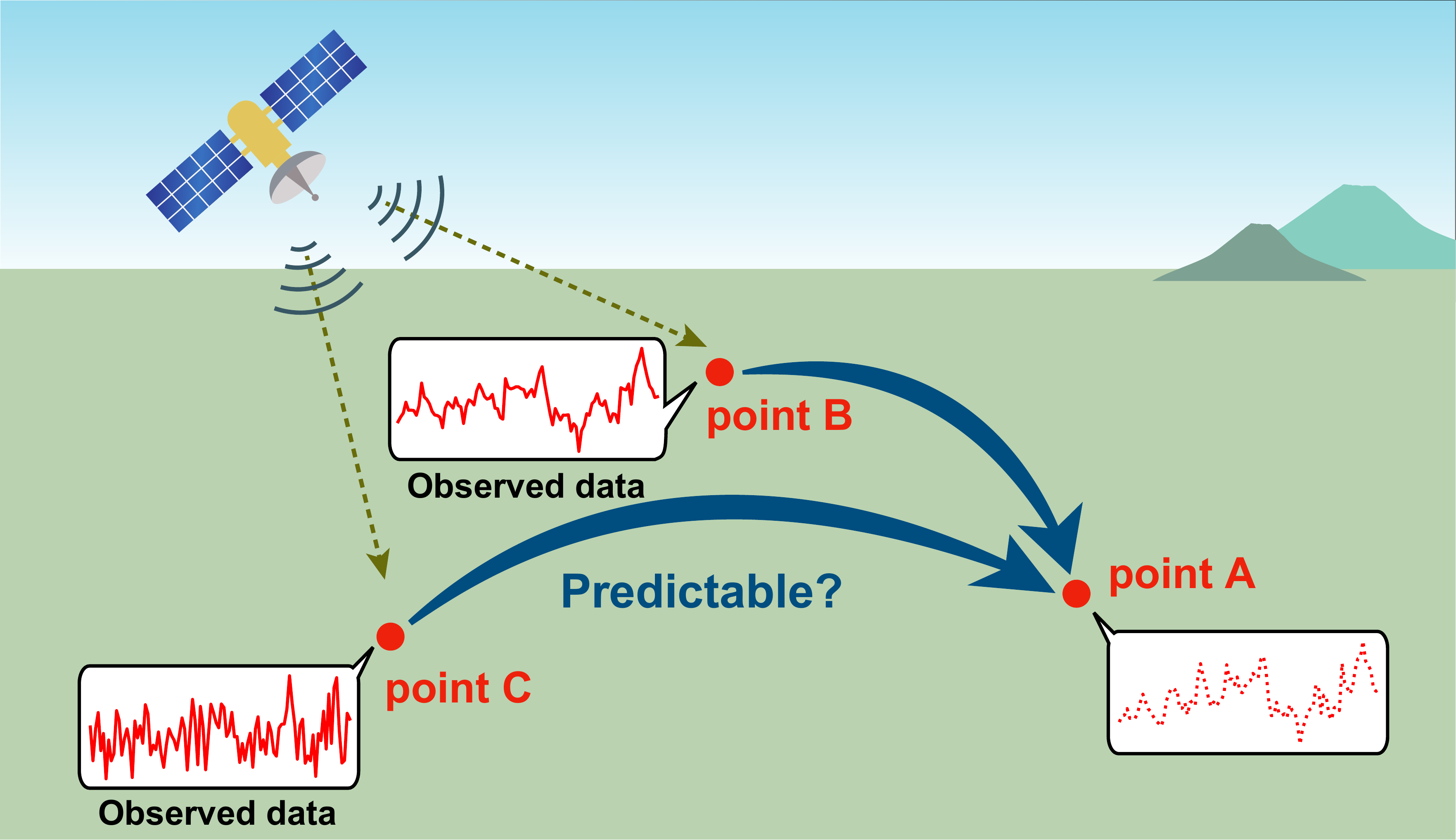}
 \caption{Conceptual diagram of time series prediction in a geographically distant point. The aim is to predict ``unobserved'' climate time series data at a target location from the ``observable'' time series data. See Sec.~\ref{sec:methods} for details of prediction methods. 
 }\label{fig:concept}
\end{figure}

In this study, we deal with the problem of predicting a physical quantity related to climate change from an observed quantity at a remote location, and scrutinize the conditions for predictability through a systematic investigation of the relation between prediction accuracy and geographical distance between the two locations. The problem setup is shown in Fig.~\ref{fig:concept}. We consider the task of predicting unobserved quantity at a target remote point (point A) by using observed data at a certain observation point (point B or C), specifically focusing on near-surface air temperature and air pressure data.
As prediction methods, we use reservoir computing (RC)~\cite{MNM2002,JH2004}, a computationally efficient machine learning method for time-series data, and the vector autoregression model (VAR), a typical statistical method for analyzing time-series data. An RC model feeds input data into a system with nonlinear dynamics, called a reservoir, and performs a pattern analysis for the reservoir state in a linear readout layer. Compared to deep learning models, the computational cost for model training is highly reduced in RC by restricting the trainable parameters in the readout layer~\cite{JH2004}. A VAR model is an autoregressive model that considers the interaction between multiple time series variables and represents the current value of the variables as a linear function of the past values of them. Compared to conventional machine learning methods, these methods do not require fine-tuning of the model to suit the target, and the learning time is very fast, making them suitable for online data processing.

We perform numerical experiments by using reanalysis data provided by the Japan Meteorological Agency 55-year long-term reanalysis project (JRA-55). The results show that there is a relation between the distance from observation to target points and prediction performance, and that it is necessary to use data from observation points reasonably close to the target point in order to make accurate predictions.
We also influence the impact of input-output data correlation on prediction performance. Although it is intuitively plausible that the data correlation and the prediction performance are related, the quantitative relation between them has not yet been fully investigated. 
Our quantitative analysis of the relation between data correlation and distance-dependent predictability reveals the extent to which the estimation of unobservable climatic elements at distant points is valid. 

Furthermore, we find that the RC model outperforms the VAR model for tested data with a particularly high correlation in the predictive domain.
This indicates that the RC approach can effectively extract nonlinear features of the input data by exploiting a nonlinear transformation in the reservoir.
Our results also provide quantitative insight into the performance and limitations of machine learning-based methods for predicting climate data, which will become increasingly important in AI-based meteorology.
In particular, the relation between the correlation of the analyzed input-output data and the prediction performance is expected to be useful in selecting the data observation points in the future.
%
%======
%
\section{\label{sec:results}Results}
\begin{figure*}[htb]
 \centering
 \includegraphics[keepaspectratio, width=1.7\columnwidth] {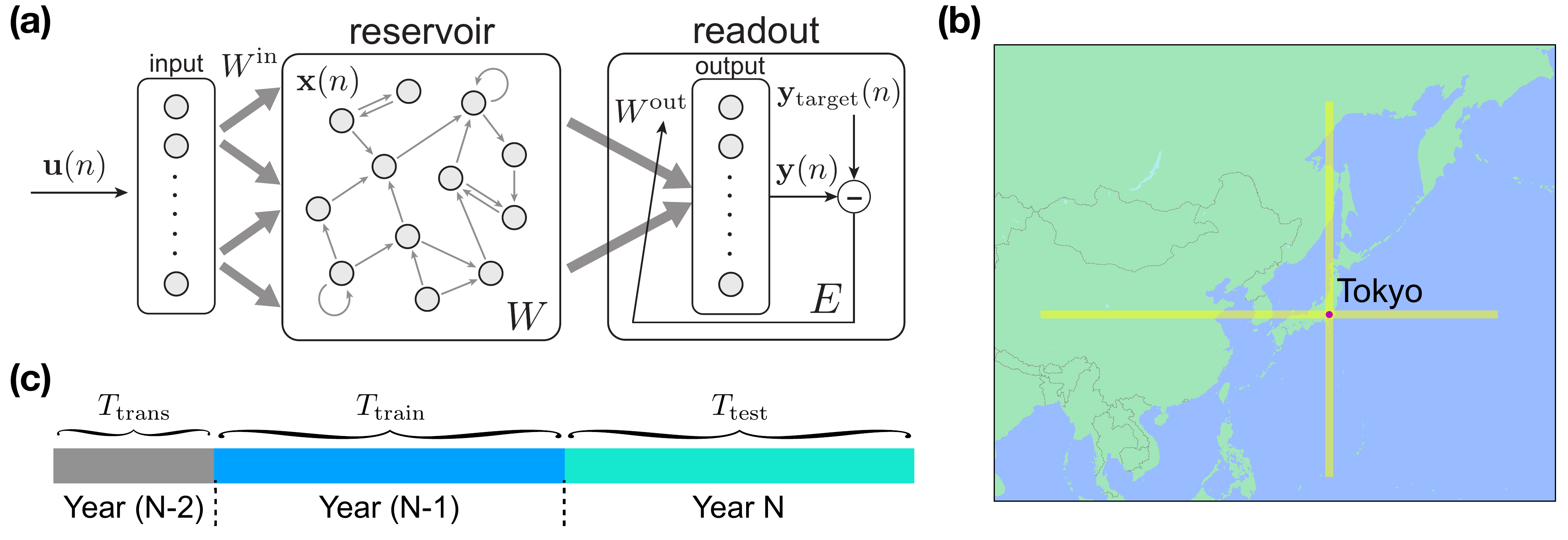}
  \caption{
  (a) An RC model called Echo State Network~\cite{Jaeger2001}. (b) Location of the observation point and the target point. The observation point is one point on a yellow band, and the target point is the cross point of two yellow bands, indicated as Tokyo in this figure. (c) Overview of the Year $N$ dataset. The transient data with length $T_\mathrm{trans}$ is the last 300 steps of Year $(N-2)$ data, the training data with length $T_\mathrm{train}$ is the whole of Year $(N-1)$ data, and the test data with length $T_\mathrm{test}$ is the whole of Year $N$ data.}
 \label{fig:methods}
\end{figure*}

\begin{figure*}[htb]
 \centering
 \includegraphics[keepaspectratio, width=1.7\columnwidth] {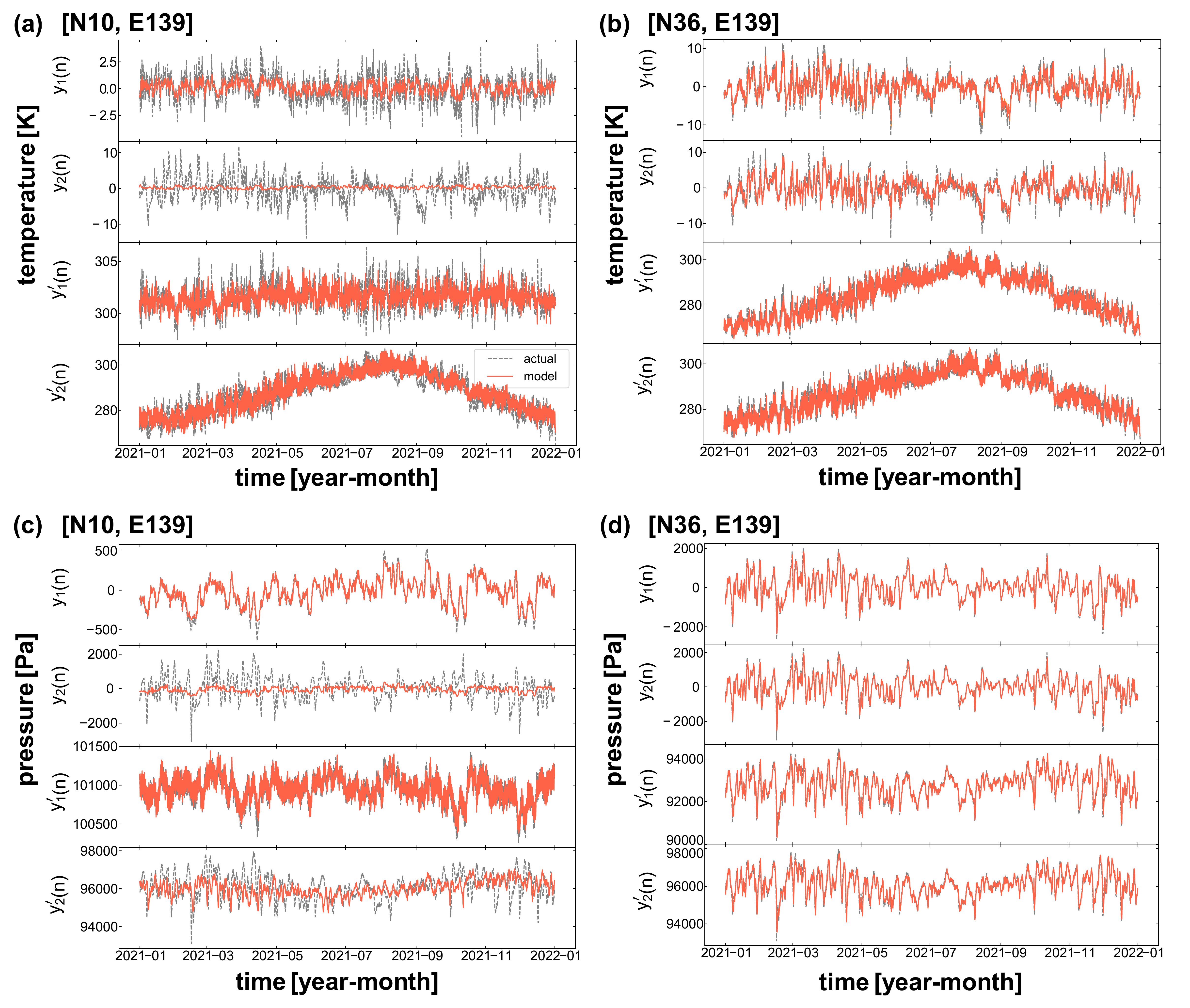}
  \caption{
  Results of RC-based prediction showing Year 2021 data. (a),(b) temperature, (c),(d) pressure. Predicted time course (red solid line) is superimposed on the test data. The time series data after removing the seasonal trend are used for training. The variables $y_1(n)$ and $y_2(n)$ denote the actual input and output data, respectively, while $y_1'(n)$ and $y_2'(n)$ denote input and output data restored to original scale, respectively.}
 \label{fig:results_esn}
\end{figure*}

In this study, we predict unobserved climate time series data $y_2(n)$ at a distant point (point A in Fig.~\ref{fig:concept}) from observed time series data $y_1(n)$ of the same climate variable at a point (point B or C in Fig.~\ref{fig:concept}) by using a representative RC model called Echo State Network (ESN) shown in Fig.~\ref{fig:methods} (a). In this paper, we refer to the point with observed data $y_1(n)$ as the ``observation point'' and the point with unobserved prediction data $y_2(n)$ as the ``target point''. The JRA-55 reanalysis data are used as the dataset (see Sec.~\ref{sec:dataset}). The input to the ESN is the data at a point located in the east/west or south/north of the target point as shown with yellow bands in Fig.~\ref{fig:methods} (b). As shown in Fig.~\ref{fig:methods} (c), the time series data at each location extracted from the JRA-55 dataset are divided into the transient, training, and test periods for the ESN-based prediction. We eliminate the seasonal trend by using the difference data from the historical average (see Sec.~\ref{sec:methods} for details).  
In this section, we focus on Tokyo as the target point, but we conduct similar experiments with other target points including New York, London, Cairo, and Canberra. The results for these cities are discussed in the Sec.~\ref{sec:discussion}.
Typical prediction results are shown in Fig.~\ref{fig:results_esn} for the cases with the observation point (N10,E139) far away from and the other point (N36,E139) near the target point. These results are for Year 2021 dataset. Figures~\ref{fig:results_esn} (a)-(b) show the results for temperature prediction and Figs.~\ref{fig:results_esn} (c)-(d) show those for pressure prediction. Each figure shows the predictions of the difference data $y_1(n), y_2(n)$ and the corresponding values $y'_1(n), y'_2(n)$ restored to the original scale by adding the historical average to them.
Defining ${\bf{y}}(n)=(y_1(n), y_2(n))^\top$, we can write ${\bf{y}}'(n) :={\bf{y}}(n)+ {\bf{y}}_\mathrm{ave}$, where ${\bf{y}}_\mathrm{ave}$ is the historical average vector. The training data corresponds to the period from Jan. 1, 2020 to Dec. 31, 2020 (not shown). The prediction results indicate the test period from Jan. 1, 2021 to Dec. 31, 2021 (red solid line), superimposed on the actual data (i.e. correct data, gray dashed line). When the observation points are near the target point, the prediction results generally follow the actual data for both $y_1, y_2$, meaning that the predictions are highly accurate. On the other hand, when the observation point is far away from the target point, the predicted values of the difference data $y_2$ oscillate near zero, resulting in unsuccessful prediction. When restored to the original scale, especially in the case of temperature, the fluctuating behavior is well reproduced with the trained ESN model. Note, however, that this is because the prediction error in the difference data is much smaller than that in the original scale data and then the prediction result oscillating near zero is almost equivalent to using the historical average value, subtracted when creating the difference data, as the prediction.

\begin{figure*}[hbt]
  \centering
  \includegraphics[keepaspectratio, width=1.7\columnwidth]
  {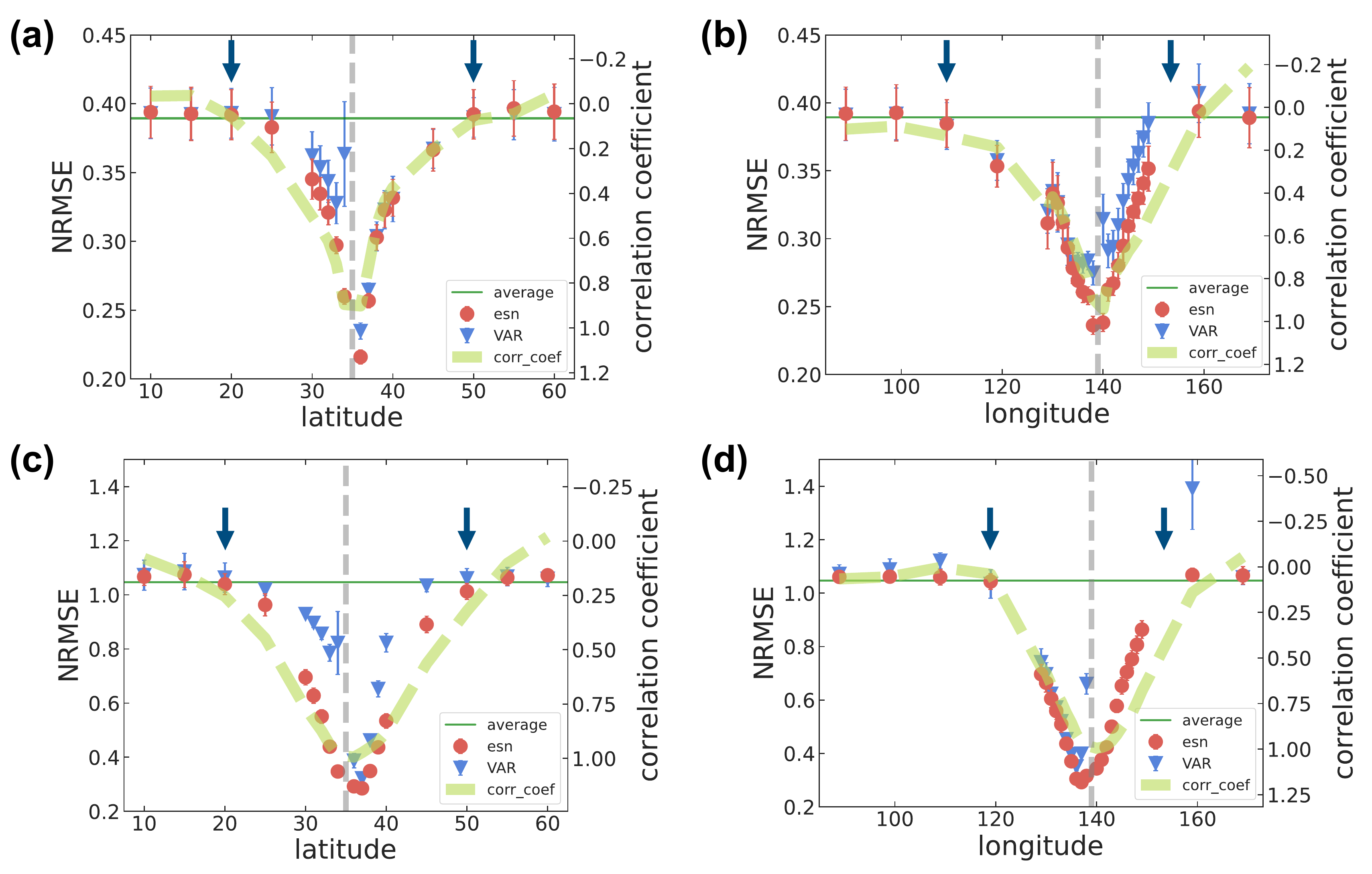}
 \caption{
 Performance comparison of prediction methods in NRMSE. The prediction methods include RC-based prediction (red filled circles), VAR-based prediction (blue downward triangles), and historical average based prefiction (green solid line). (a),(b) temperature, (c),(d) pressure prediction. The yellow dashed lines are the scaled correlation coefficients between true time series data at the observation and target points. Vertical dashed lines indicate the location of the target point (Tokyo). The downward arrows show the thresholds where the NRMSE reaches a plateau.}\label{fig:comparison}
\end{figure*}
In this study, prediction performance is evaluated on the original-scale data ${\bf{y}}'(n)$ by using the following normalized root mean squared error (NRMSE):
\begin{align}
\mathrm{NRMSE} := \frac{\sqrt{\langle \|({\bf{y}}'_\mathrm{target} (n) - {\bf{y}}'(n)\|_2^2\rangle_\mathrm{test}}}{\sqrt{\langle\|{\bf{y}}'_\mathrm{target} (n) - \langle {\bf{y}}'_\mathrm{target}(n)\rangle_\mathrm{test}\|_2^2\rangle_\mathrm{test}}},
\end{align}
where $\langle \cdot \rangle_\mathrm{test}$ represents the average over the test period of length $T_\mathrm{test}$.
Figure~\ref{fig:comparison} shows the NRMSE as a function of the latitude or longitude of the observation point. The NRMSEs are calculated for five time series data from Year 2017 to 2021. The marker indicates their mean, and the error bar indicates their variance. The vertical dashed line indicates the location of the target point, i.e. Tokyo at (N35, E139) in this figure. In all the panels, the NRMSEs are quite small when the observation points is near the target point, and the NRMSE increases with the distance from the target point to the observation point. Moreover, the rate of increase in the NRMSE reaches a plateau when the distance from the target point exceeds a certain threshold indicated by downward arrows in Fig.~\ref{fig:comparison}. For example in Fig.~\ref{fig:comparison} (a), this threshold is roughly 15 degrees from the target point. Compared with the NRMSE obtained when the historical average value is used as a prediction, indicated by the green solid line in Fig.~\ref{fig:comparison}, the RC-based predictions and the historical average based prediction almost match with each other around this threshold. Hence, in this paper, we suppose that the RC has the ability to make predictions effectively within a range that does not exceed this threshold. Defining this range as the RC predictable range, for both temperature and pressure prediction, the predictable range is about 15 degrees north and south from the target point in the latitudinal direction. On the other hand, there is asymmetry in the predictable range in the longitudinal direction for both temperature and pressure, which is about 20-30 degrees west and 15-20 degrees east from the target point. This asymmetry is probably caused by the westerly winds which transport climatic substance from west to east, which requires further investigation beyond this study.

In addition, for the same time series data, we also perform predictions by using the VAR model, which is one of the standard multivariate models for statistical time series analysis. The NRMSEs for the VAR-based predictions are shown in Fig.~\ref{fig:comparison} as blue down-triangle plots. As the RC-based predictions, the VAR model predictions show an increase in the NRMSE with the distance between the observation point and the target point. The rate of increase in the NRMSE saturates at approximately the same threshold as that in RC predictions. However, for almost all observation points, RC predictions are more accurate than VAR-based ones in the predictable range, i.e. the range where the NRMSEs are smaller than those obtained using the historical average. In temperature prediction, the NRMSE averaged over the neighboring region within $\pm 10^\circ$ from the prediction point is smaller with the RC-based method than with the VAR-based method, indicating that the RC model achieves more effective prediction than the VAR model. We confirmed that this consequence is also true for the other data treated in Sec.~\ref{sec:discussion}.
At observation points sufficiently far away from the target point, the prediction error of the VAR model is close to that of the RC model and the historical average. In this situation, it is enough to simply use the historical average as the predicted value in terms of computational cost. We also notice that in the VAR model predictions, there are some cases where the NRMSE is abnormally large even when the observation point is very close to the target point, e.g. the observation point at (N34, E139) in Fig.~\ref{fig:comparison} (a). This phenomenon is attributed to a failure in the estimation of the coefficient matrix, givin by ${\bf{\Phi}}_1$ in Eq.~\eqref{eq:VAR}, of the VAR model due to high data correlation. This result indicates that stable prediction with the VAR model is difficult when the time series data correlation is very high.
As in Fig.~\ref{fig:comparison} (d), the NRMSEs of the VAR model are missing in the range of [E140, E150], because their values diverge and predictions are virtually impossible throughout the test period.

Such a problem is caused by the condition that the VAR-based and RC-based predictions use the same known information for fair comparison. This may be avoided by introducing constraints such that the absolute values of the coefficients do not exceed 1 during the training of the VAR model. However, if the model has ability to express the time series data of interest, it is possible to select appropriate coefficients without such constraints. Our results suggest that the VAR model has less complexity and usability in approximating climate time series data than the RC model.

Furthermore, we examine the correlation coefficients between the true time series data $y_1$ at the observation point and $y_2$ at the target point. The correlation coefficients are calculated for the test period of the five time series data from Year 2017 to 2021, and their mean value is obtained for each observation point. The values of the correlation coefficients are larger/smaller in the regions where the NRMSEs by both RC and VAR models are small/large, hence these two factors appear to be correlated. This correlation is obvious from the fitted line for the scatter plots in Fig.~\ref{fig:discussion}. The curves of the scaled correlation coefficients shown as yellow dashed lines in Fig.~\ref{fig:comparison}, roughly follow the NRMSEs of the RC prediction. This result is consistent with our intuition that the correlation between observation and target time series data is an important factor that characterizes the prediction performance.
%
%======
%
\section{\label{sec:discussion}Discussion}
To validate our results, we performed similar prediction tasks using RC and VAR models by changing the target point from Tokyo to New York, London, Canberra, and Cairo (See Supplemental Materials for details). For all cities, as in the case of Tokyo, the prediction error increases with the distance between observation and target points, and the NRMSEs are smaller with the RC model than with the VAR model for almost all observation points. The east-west asymmetry shown in Fig.~\ref{fig:comparison} are also reproduced. The VAR model tends to have difficulty in making a prediction from data obtained in the vicinity of the target location as in the case with Tokyo. On the other hand, there is no strong asymmetry in the correlation coefficients between the true time series data at the observation point and that at the target point. The asymmetry property in the prediction results could be attributed to whether the data at the target location is advanced or delayed compared with the data at the observed location.

\begin{figure}[tb]
  \centering
  \includegraphics[keepaspectratio, width=.95\columnwidth]
  {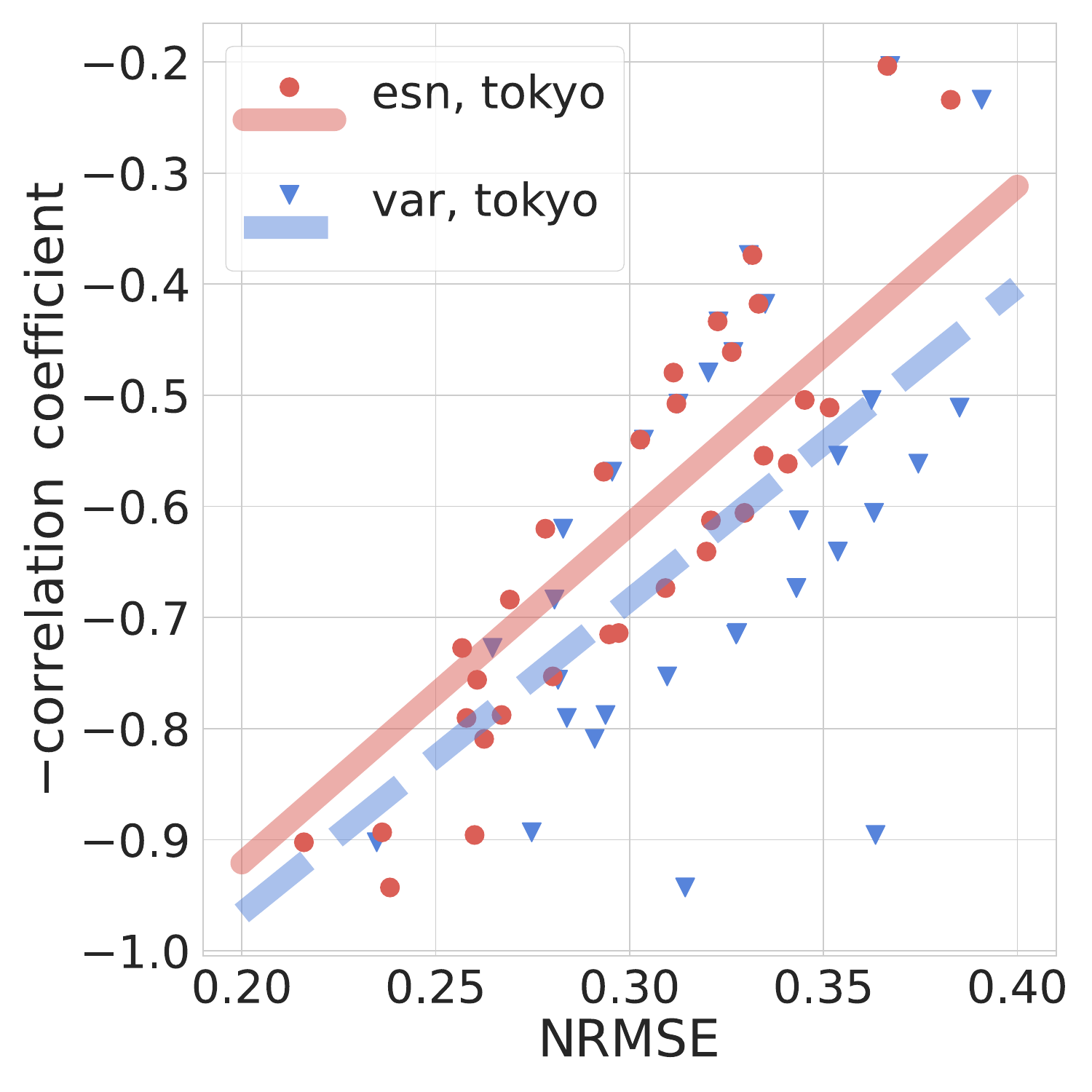}
 \caption{The relation between NRMSE and input-output correlation. The NRMSE values are obtained in temperature prediction for the target points at Tokyo (RC: red circles, VAR: blue downward triangles). The regression lines are drawn for the RC model (red solid line) and VAR model (blue dashed line). The fitting parameters of the lines are determined for the RC and VAR results assuming that the slope parameter is independent and the intercept parameter is common.}\label{fig:discussion}
\end{figure}

The relation between the prediction error (NRMSE) and observation-target time series correlation is shown in Fig.~\ref{fig:discussion} when the target point is Tokyo. Since we are interested in the range with good predictions, we show only the results for observation points less than $\pm 10^\circ$ away from the target point. We can clearly see that they are positively correlated, indicating that the similarity of climate time series data at geographically distant points is a key factor for good accuracy of the predictions. Such positive correlations are also observed when the target location is another city and/or when predicting pressure. According to these results, it is expected that the applicability of RC models in our problem can be determined by simply checking the correlation between past input/output data in advance. This finding is quite useful when we need to fully interpolate missing or uncertain data at target points from those at a minimum number of observation points. Since the climate data used in this study tend to have stronger correlations as the distance between observation points is shorter, it is possible to estimate the predictable range from the relation between the NRMSE and the distance. We can obtain the regression lines showing this relation for the RC and VAR models in the same figure. The regression lines are drawn so that the intercepts of the two lines are common. A comparison between them shows that the RC model outperforms the VAR model for data with the same correlation strength. The superiority of the RC model is most prominent when the correlation is very high ($>0.9$).

At an observation point sufficiently far away from the target point, the NRMSE approaches around the error value $\mathrm{NRMSE}_\mathrm{ave}$ with the historical average (the solid green line in Fig.~\ref{fig:comparison}). This is a baseline of the prediction error for each city: the lower the latitude, the smaller the baseline value (See Table~\ref{table:city}). In particular, Cairo has much lower NRMSE than the other cities, which might be related to the fact that only Cairo belongs to class B in the K\"{o}ppen climate classification. It is possible that climatic differences affect predictability, though a further research is needed to clarify the relation.
\begin{table*}[bt]
\caption{Baseline NRMSE values for the five target cities. The columns from left to right indicate the following items: target city, baseline error values for temperature, those for pressure, latitude, longitude, and the climate group in the K\"{o}ppen climate classification.}\label{table:city}
\centering
\begin{tabular}{lccccc}\hline
Target city & $\mathrm{NRMSE}_\mathrm{ave,temp}$ & $\mathrm{NRMSE}_\mathrm{ave,pres}$ & Latitude & Longitude & Climate \\ \hline\hline
Cairo & 0.3599 & 0.7732 & 30.1N & 31.4E & Bwh\\
Tokyo & 0.3894 & 1.047 & 35.7N & 139.7E & Cfa \\ 
NewYork & 0.5060 & 1.125 & 40.8N & 73.9W & Cfa \\
London & 0.5942 & 1.153 & 51.5N & 0.1W & Cfb \\
Canberra & 0.5794 & 1.083 & 35.3S & 149.2E & Cfb \\
\hline
\end{tabular}
\end{table*}
\section{Conclusions}
In this study, we used the RC framework to predict the ``unobserved'' climate quantity at a remote location from the same quantity at an observation point. The results have revealed that good prediction results are obtained when the input points are reasonably close to the target points, suggesting that the prediction performance is characterized by the degree of correlation between the observed and target data. This approach enables to pre-select datasets for which the RC model works well. For the climate data we have treated in this study, it is intuitively clear that the observation-target distance and the time series data correlation are closely related, although the actual correlation can be examined only when both data in the past are available at hand. Our results support the intuition quantitatively with the estimation of predictable ranges for the five target cities.
In this study, we focused on the same physical quantity (temperature or pressure), as the observable and the target. A possible extension of our work is to apply timescale-rich RC-based methods~\cite{TMY+2022} to predicting different physical quantities at the same location or at a distant location.
The application of RC-based prediction of unobserved elements is not limited to climate data, but can be extended to other nonlinear time series data, such as biological and financial ones. The concept of input-output time series correlation presented in this study is a simple but effective indicator associated with predictability of data-driven approaches and a useful measure for selecting appropriate information to be used.
%
%======
%
%TC:ignore
\section{\label{sec:methods}Methods} 
\subsection{\label{sec:reservoir}Reservoir computing (RC)}
In this study, we predict climate time series data by using the standard Reservoir Computing model, called Echo State Network (ESN)~\cite{JH2004,Jaeger2001,Gauthier2018}, illustrated in Fig.~\ref{fig:methods} (a). The ESN is a special variant of recurrent neural networks (RNNs) consisting of an input layer, a reservoir, and a readout. It maps time series input data into a high-dimensional feature space using the reservoir, and then reads out the features of the input data through its dynamical behavior in the reservoir from the readout. The ESN model can be trained with very low computational cost compared to the fully trainable RNN because it learns only the output weights $W^\mathrm{out}$ in the readout. In the ESN, the recurrent network with fixed connection weights acts as a reservoir that generates ``echoes'' of the past input via the feedback mechanism. Physical reservoir computing, in which nonlinear transformation functions of reservoirs are realized by real-world physical phenomena, has also attracted much attention recently~\cite{TYH+2019,Nakajima2020,BL2021,RAC+2022}.

We denote the input vector at time $n$ by ${\bf{u}}(n) \in \mathbb{R}^{N_u}$. The time evolution of the reservoir state vector ${\bf{x}}(n) \in \mathbb{R}^{N_x}$ and the output vector ${\bf{y}}(n) \in \mathbb{R}^{N_y}$ are written as follows:
\begin{align}
\label{eq:esn1}
&{\bf{x}}(n) = f(W^\mathrm{in}{\bf{u}}(n) + W{\bf{x}}(n-1)), \\
\label{eq:esn2}
&{\bf{y}}(n) = W^\mathrm{out}{\bf{x}}(n),
\end{align}
where $N_u,N_x (\gg N_u),N_y$ are the number of nodes in the input, reservoir and output layers, respectively, and $W^\mathrm{in},W,W^\mathrm{out}$ are the weight matrices between the input and reservoir layers, within the reservoir, and between the reservoir and output layers, respectively. The function $f(\cdot)$ denotes the element-wise activation function, which is set as $f=\tanh$ in this study. The ridge regression is employed as a learning algorithm, and the optimal output weight matrix $\hat{W}^\mathrm{out}$ is obtained by minimizing the error function $E(W^\mathrm{out})$ as follows:
\begin{align}
&\hat{W}^\mathrm{out} = \mathrm{argmin}_{W^\mathrm{out}} \left[E(W^\mathrm{out}) + \beta\| W^\mathrm{out}\|_F^2 \right] ,\label{eq:objective-func}\\
&E(W^\mathrm{out}) = \frac{1}{T}\sum_{n=1}^T \|{\bf{y}}(n) - {\bf{y}}_\mathrm{target}(n)\|_2^2,
\end{align}
where $\| \cdot \|_F$ denotes Frobenius norm.

\subsection{\label{subsec:var} Vector autoregression (VAR) model}
The VAR model is a standard statistical model for multivariate time series analysis.
We compare the RC prediction results with the VAR prediction results.
The VAR model describes the time evolution of vector ${\bf{y}}(n) \in \mathbb{R}^{N_v}$, where
${\bf{y}}(n)$ is the set of state variables at the $n$th time step and $N_v$ is the dimension of the vector. The time evolution of ${\bf{y}}(n)$ is written as follows:
\begin{align}
\label{eq:VAR}
{\bf{y}}(n) = \,&{\bf{\Phi}}_1{\bf{y}}(n-1) + {\bf{\Phi}}_2{\bf{y}}(n-2) + \cdots \nonumber \\
&+ {\bf{\Phi}}_p{\bf{y}}(n-p) + {\bf{c}} + {\bm{\varepsilon}}(n),
\end{align}
where ${\bf{\Phi}}_k$ is a coefficient matrix of the $k$-step-behind state ${\bf{y}}(n-k)$, ${\bf{c}}$ is a constant vector, ${\bm{\varepsilon}}(n)$ is a noise vector, and $p$ is the order of the VAR model. In this study, we set $p=1$ and $N_v=2$.
For comparison with RC, the time series data are divided into three time periods, $T_\mathrm{trans}, T_\mathrm{train}, T_\mathrm{test}$, each of which have the same length as that in the RC setup, but the transient period $T_\mathrm{trans}$, which is necessary for the RC, is not used in the VAR model. 
In the RC-based prediction, the unobserved data $y_2$ are predicted from only the observed data $y_1$. 
To make a fair comparison with the RC prediction, the regressions are performed for the training period of length $T_\mathrm{train}$ to determine the coefficient matrix ${\bf{\Phi}}_1$ and the constant vector $\bf{c}$.
In the test period, real (i.e. observed) data are used only as $y_1(n-1)$ to predict $y_2(n)$ with the fixed regression coefficients. 
Note that in this procedure $y_2(n-1)$ is the 1-step earlier predicted value since $y_2$ is the unobserved data through the test period. Therefore, by representing the observed data at the observation point as $y_{1,\mathrm{observed}} (n)$, the predicted time series data at the target point for the test period are obtained as follows: 
\begin{align}
\label{eq:VAR_test2}
y_2(n) = \phi_{21} y_{1,\mathrm{observed}} (n-1) + \phi_{22} y_2 (n-1) + c_2,
\end{align}
where $\phi_{ij}$ and $c_i$, for $i,j\in\{1,2\}$ are the components of ${\bf{\Phi}}_1$ and ${\bf{c}}$, respectively. For simplicity, the noise term in Eq.~\eqref{eq:VAR} is set to ${\bm{\varepsilon}}(n)=(0,0)^\top$. We use the python package statsmodels~\cite{statsmodels} in the VAR-based prediction to obtain ${\bf{\Phi}}_1$ and ${\bf{c}}$.

The VAR model is trained so that the absolute value of the eigenvalues of the coefficient matrix is less than 1, but the individual coefficients are not necessarily less than 1. Since $y_2$ is the 1-step earlier predicted value, the predicted value of $y_2$ diverges when the absolute value of the coefficient of $y_2$ is larger than 1 in the prediction using only Eq.~\eqref{eq:VAR_test2}.

\subsection{\label{sec:dataset}Dataset}
This subsection describes the dataset. In this study, we use the long-term reanalysis data JRA-55 provided by the Japan Meteorological Agency 55-year long-term reanalysis project~\cite{jra-55}. The datasets of JRA-55 6-Hourly Model Resolution Surface Analysis Fields obtained from the NCAR website~\cite{ncar} are used.

The JRA-55 data are provided in the GRIB format. We first process the data in the GRIB format to create time series data for each location, which will be the input and output for the prediction models. The data at the target point are obtained by averaging the values of a climatic element inside the 1 degree of mesh including the target point. For example, the data at the point of (N35, E139) adopt the average value inside of closed interval [N35, N36] and [E139, E140]. More precisely, we select the closest mesh boundary for the given latitude/longitude and obtain the average value within these boundaries since the meshes of the original dataset are not cut every 1 degree. The length of each time series data is 3220 steps every six hours, namely 2 years and 2.5 months in real time, and five sets of data with different periods are created for each location. In order to align the lengths of all time series data, the data for February 29 are excluded in the case of leap years. In addition, previous studies indicate that time series data can pose difficulties in generating effective predictors when they have some kinds of trends~\cite{LB2011,SBA+2012}. An effective approach to avoiding this situation is a technique called seasonal decomposition, which typically decomposes the time series data into trend-cycle, seasonal, and irregularity components~\cite{WS2010}. We create difference data by subtracting the average value of the past 3 years from the original data, herein referred to simply as the historical average, or removing the trend more simply. We use the difference data as the actual input/output in the RC/VAR prediction.

\subsection{\label{sec:problem}Problem setting}
In this study, we discuss the predictability of climate variables by using RC. The time series data of surface temperature ``temp'' and surface pressure ``pres'' are used as climate elements. Our goal is to predict unobserved data of interest at the ``target point'' (point A in Fig.~\ref{fig:concept}) from observed data at a certain distant point away from there (point B or C in Fig.~\ref{fig:concept}, hereafter, we call this point an ``observation point'').
The procedure for predicting data at Tokyo (N35, E139) is explained as an example. The model training is performed by using two time series data. One is the data at Tokyo, and the other is the data at an observation point located east/west or north/south of Tokyo in the range [N10,N60] and [E89,E179] shown as yellow bands in Fig.~\ref{fig:methods} (b).  
In the test period, only the data at the observation point are used as input data, and the model output is the vector of ${\bf{y}}=(y_1,y_2)^\mathsf{T}$ where $y_1$ and $y_2$ correspond to the observation point and the target point (Tokyo), respectively.

In predicting a time series data by using RC, the time series data are divided into three periods, $T_\mathrm{trans}$, $T_\mathrm{train}$, and $T_\mathrm{test}$ as illustrated in Fig.~\ref{fig:methods} (c). $T_\mathrm{trans}$ is the length of a preliminary period to eliminate the influence of the initial state of the reservoir, set at $T_\mathrm{trans}=300$. The length of the training period is set at $T_\mathrm{train}=1460$, and that of the test period at $T_\mathrm{test}=1460$. This setting corresponds to the situation that the data from January 1 to December 31 of Year $(N-1)$ are treated as training data and the data from January 1 to December 31 of Year $N$ are treated as test data. We refer to this dataset as the data for Year $N$.

Several hyperparameters need to be set in the ESN; 
density $d$ which determines the degree of coupling of the reservoir nodes, the input scaling parameter $\gamma$ which is relevant to $W^\mathrm{in}$, the spectral radius $\rho$ defined as $\max(|\mathrm{eig} (W)|)$, and the regularization parameter $\beta$ for the optimization of $W^\mathrm{out}$. These parameters are set to $(d,\gamma,\rho,\beta)=(0.02,0.2,0.5,1.0)$ for temperature prediction, and $(d,\gamma,\rho,\beta)=(0.07,0.05,0.2,0.15)$ for pressure prediction. These values minimizing the average NRMSE are obtained by a grid search with a total of 40,000 parameter combinations for 15 time series data at points (N36,E139), (N25,E139), and (N10,E139) for 5 years. We fix the number of nodes in the reservoir at $N_x=400$.
%
%======
%
\section{\label{sec:acknowledgment}Acknowledgment}
This work was supported by Moonshot R\&D Grant Number JPMJMS2021, AMED under Grant Number JP23dm0307009, the International Research Center for Neurointelligence (WPI-IRCN) at The University of Tokyo Institutes for Advanced Study (UTIAS), JSPS KAKENHI (Grant Number JP20H05921, 23K28154), Council for Science, Technology and Innovation(CSTI), Cross-ministerial Strategic Innovation Promotion Program (SIP), the 3rd period of SIP ``Smart energy management system'' Grant Number JPJ012207 (Funding agency:JST), and Intelligent Mobility Society Design, Social Cooperation Program (Next Generation Artificial Intelligence Research Center, the University of Tokyo, and Toyota Central R\&D Labs, Inc.).

\section*{Data Availability}
The data that we used in this study are publicly available at \url{https://github.com/ToyotaCRDL/climate_data.git}.
%TC:endignore
% \newpage
\appendix
\section{Predictions for various cities}
\begin{figure*}[ht]
  \centering
  \includegraphics[keepaspectratio, width=1.8\columnwidth]
  {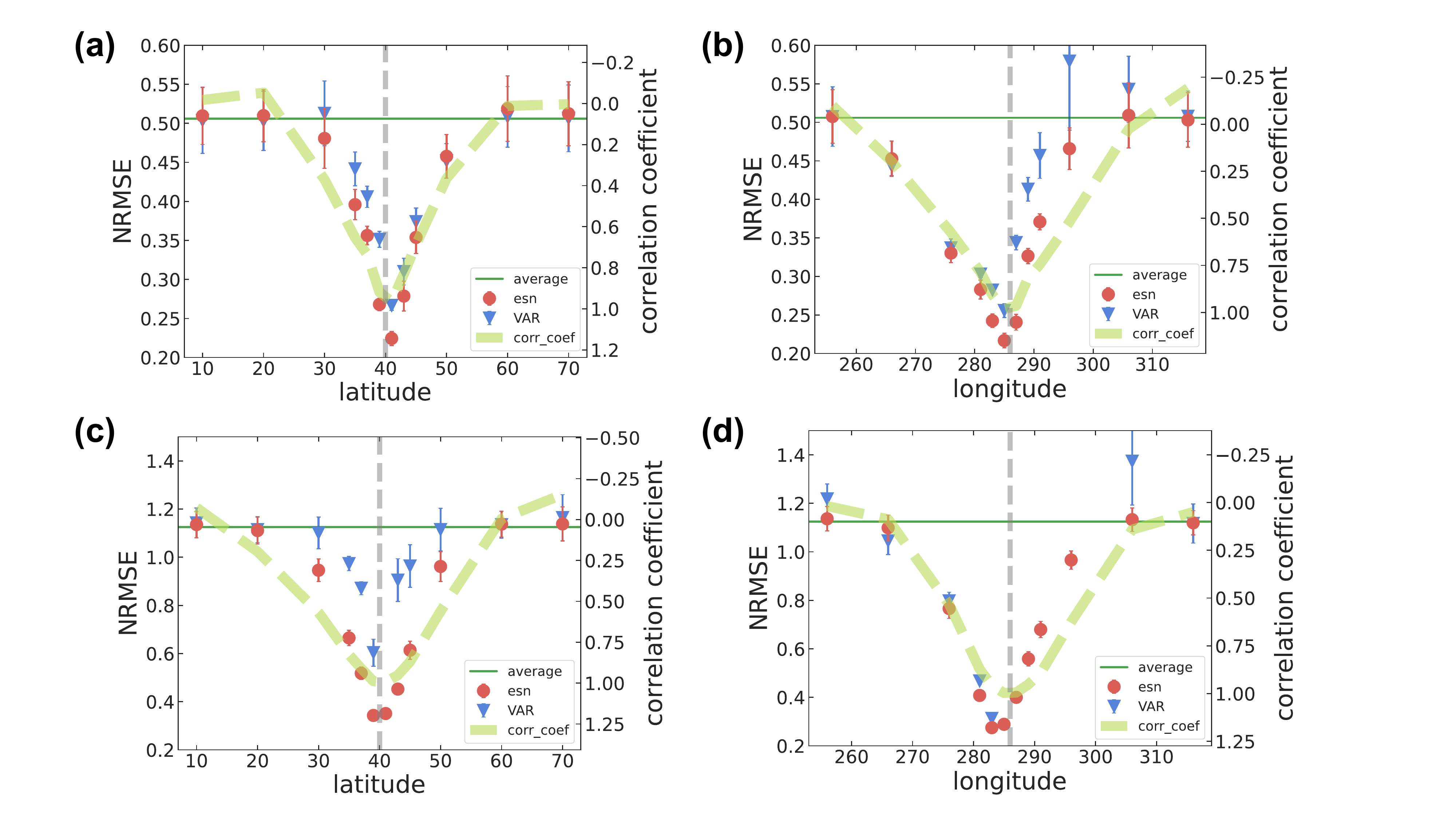}
  \vspace{-5pt}
 \caption{Performance comparison of prediction methods in NRMSE. Results for the case where target is New York. The prediction methods include RC-based prediction (red filled circles), VAR-based prediction (blue downward triangles), and histrical average based prediction (green solid line). (a) temperature, latitudinal. (b) temperature, longitudinal. (c) pressure, latitudinal. (d) pressure, longitudinal. The yellow dashed lines are the scaled correlation coefficients between true time series data at the observation and target points. The vertical gray dashed lines indicate the location of the target point.}\label{fig:result_newyork}
\end{figure*}

\begin{figure*}[hbt]
\vspace{-7pt}
  \centering
  \includegraphics[keepaspectratio, width=1.8\columnwidth]
  {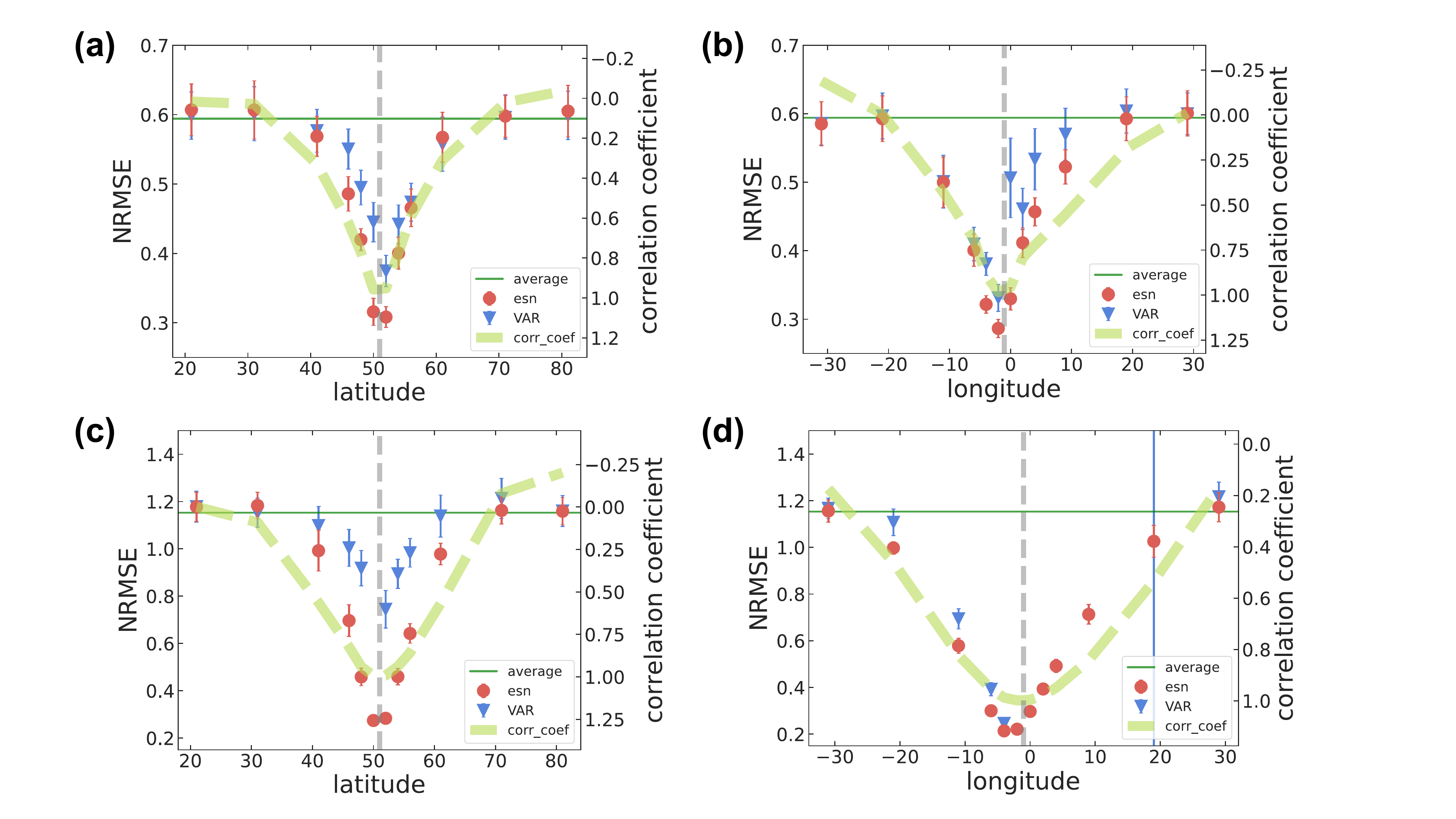}
  \vspace{-5pt}
 \caption{Results for the case where the target city is London. The quantities shown in this figure are the same as those in Fig.~\ref{fig:result_newyork}.}\label{fig:result_london}
 % \vspace{13pt}
\end{figure*}

\begin{figure*}[htb]
  \centering
  \includegraphics[keepaspectratio, width=1.8\columnwidth]
  {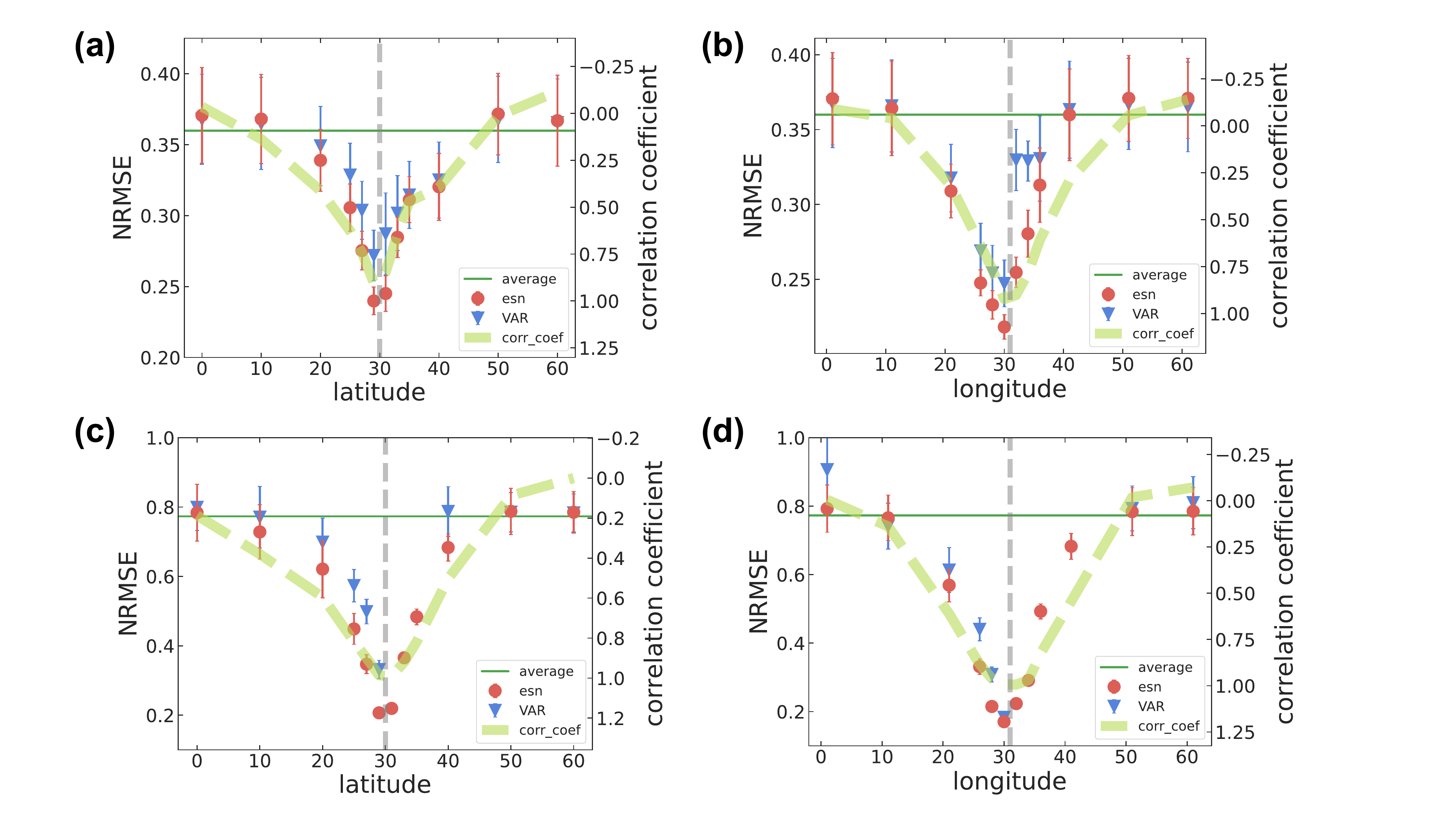}
  \vspace{-5pt}
 \caption{Results for the case where the target city is Cairo. The quantities shown in this figure are the same as those in Fig.~\ref{fig:result_newyork}.}\label{fig:result_cairo}
\end{figure*}

\begin{figure*}[hbt]
  \centering
  \includegraphics[keepaspectratio, width=1.8\columnwidth]
  {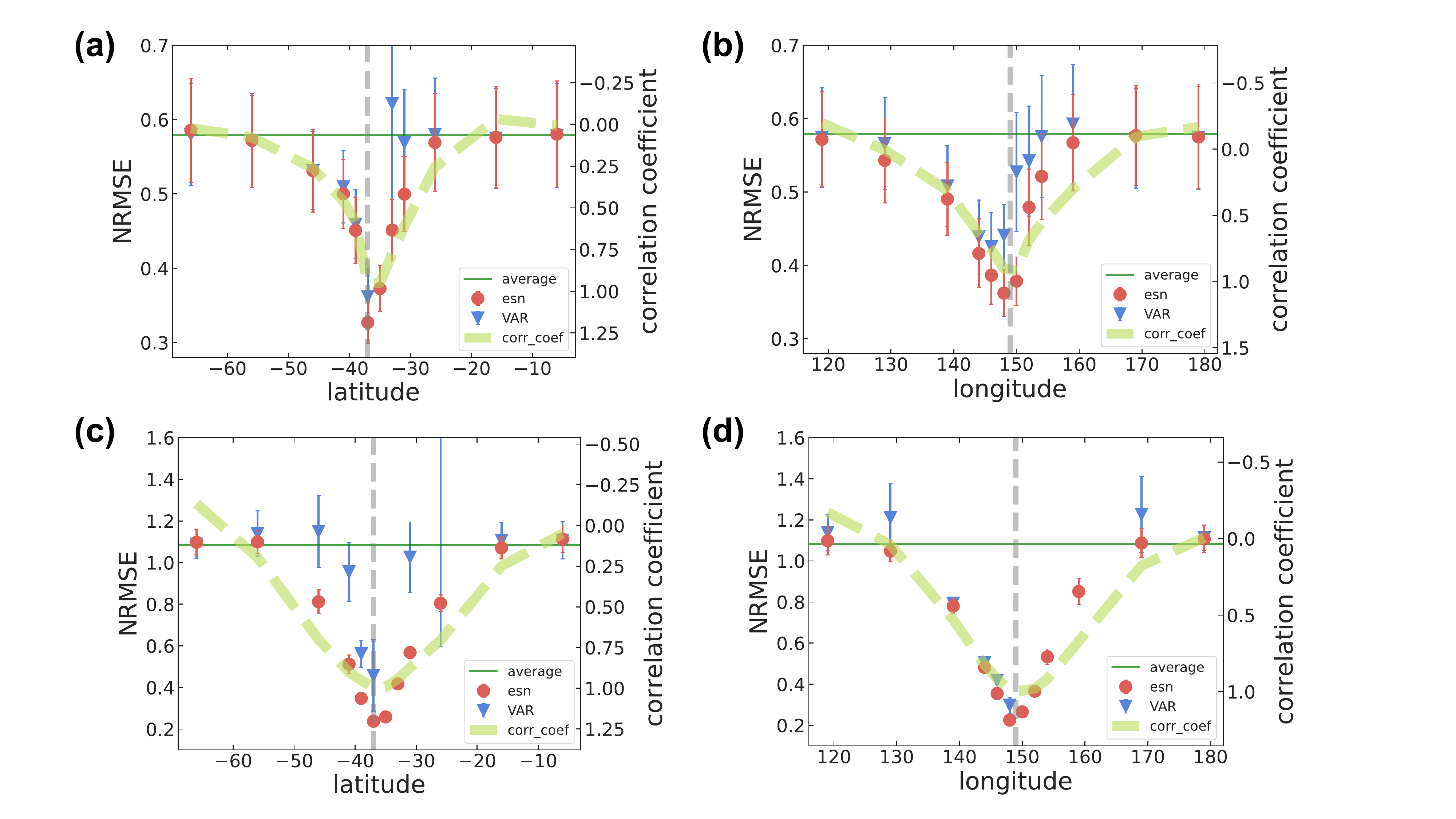}
  \vspace{-5pt}
 \caption{Results for the case where the target city is Canberra. The quantities shown in this figure are the same as those in Fig.~\ref{fig:result_newyork}.}\label{fig:result_canberra}
\end{figure*}

In addition to the case treated in the main text (Tokyo), we have also examined the RC/VAR models for the cases where the target points are New York, London, Cairo, and Canberra. The results are shown in Fig.~\ref{fig:result_newyork}-\ref{fig:result_canberra}. For the observation data, we used data at distances less than $\pm 30^\circ$ in the directions of latitude and longitude. Here, the hyperparameter values of RC are the same as those in the main text. We also made predictions using the optimized parameters in each case, but they had little effect (See Sec.~\ref{sec:parameter} in details).
\section{The reration between NRMSE and input-output correlation}
\begin{figure*}[htb]
  \centering
  \includegraphics[keepaspectratio, width=1.7\columnwidth]
  {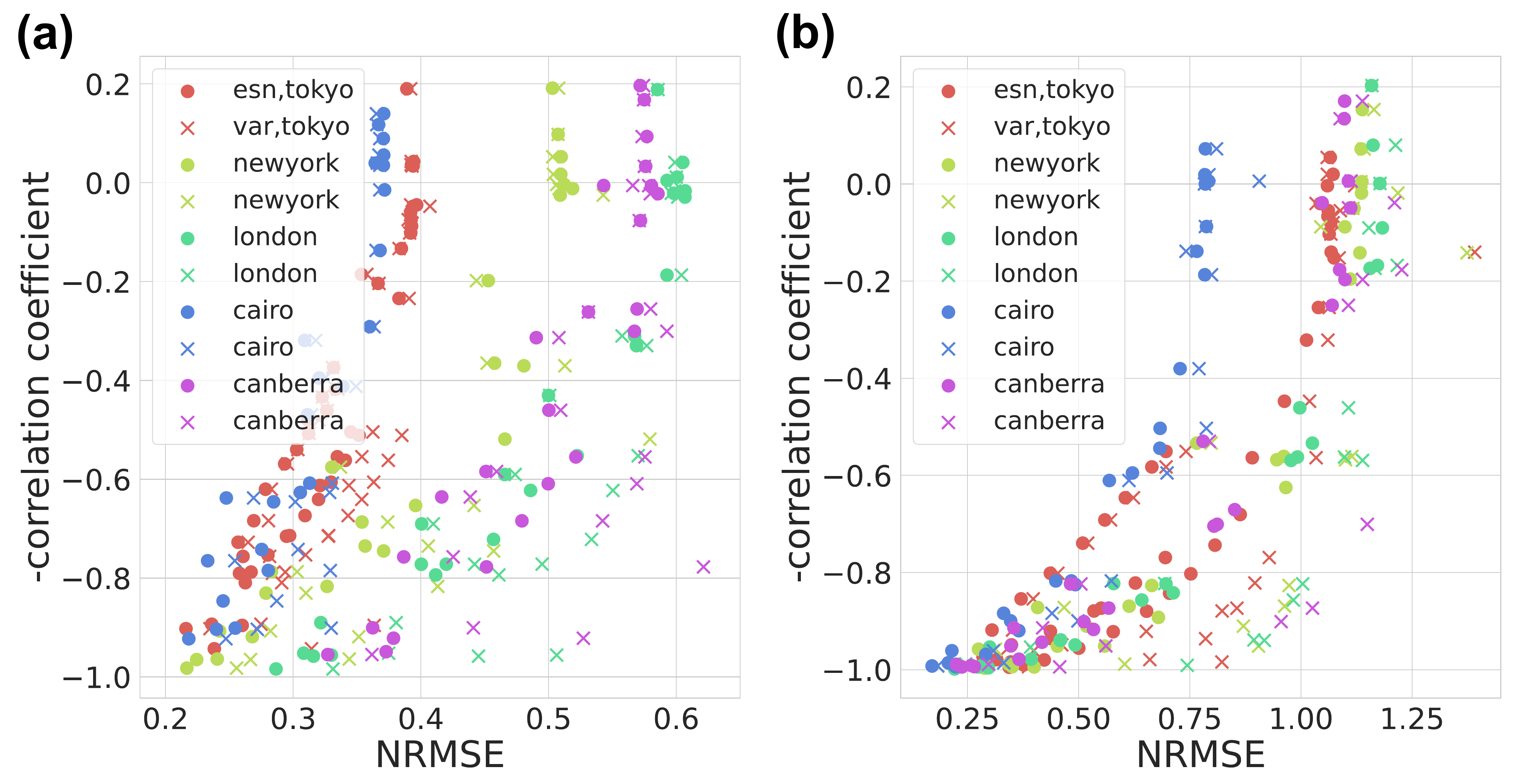}
 \caption{Relationship between NRMSE and input-output correlation coefficients. RC-based predictions (circles) and VAR-based predictions (crosses) for (a) temperature and (b) pressure are shown.}\label{fig:corr_vs_nrmse}
\end{figure*}

The relationship between NRMSE and input-output data correlation was examined for all the target cities examined in this study. Scatter plots for (a) temperature and (b) pressure are shown in Fig.~\ref{fig:corr_vs_nrmse}. The results for all the target cities show a positive correlation, even though the values differ depending on the city. Furthermore, the figure shows that the VAR tends to have a larger error than the RC in cases where the correlation is high (when the correlation coefficient $\gtrsim$ 0.6). This reflects the fact that RC is more stable when the input-output data are similar, as mentioned in the main text.
\section{Effects of the hyperparameter\label{sec:parameter}}
\begin{figure*}[htb]
  \centering
  \includegraphics[keepaspectratio, width=1.8\columnwidth]
  {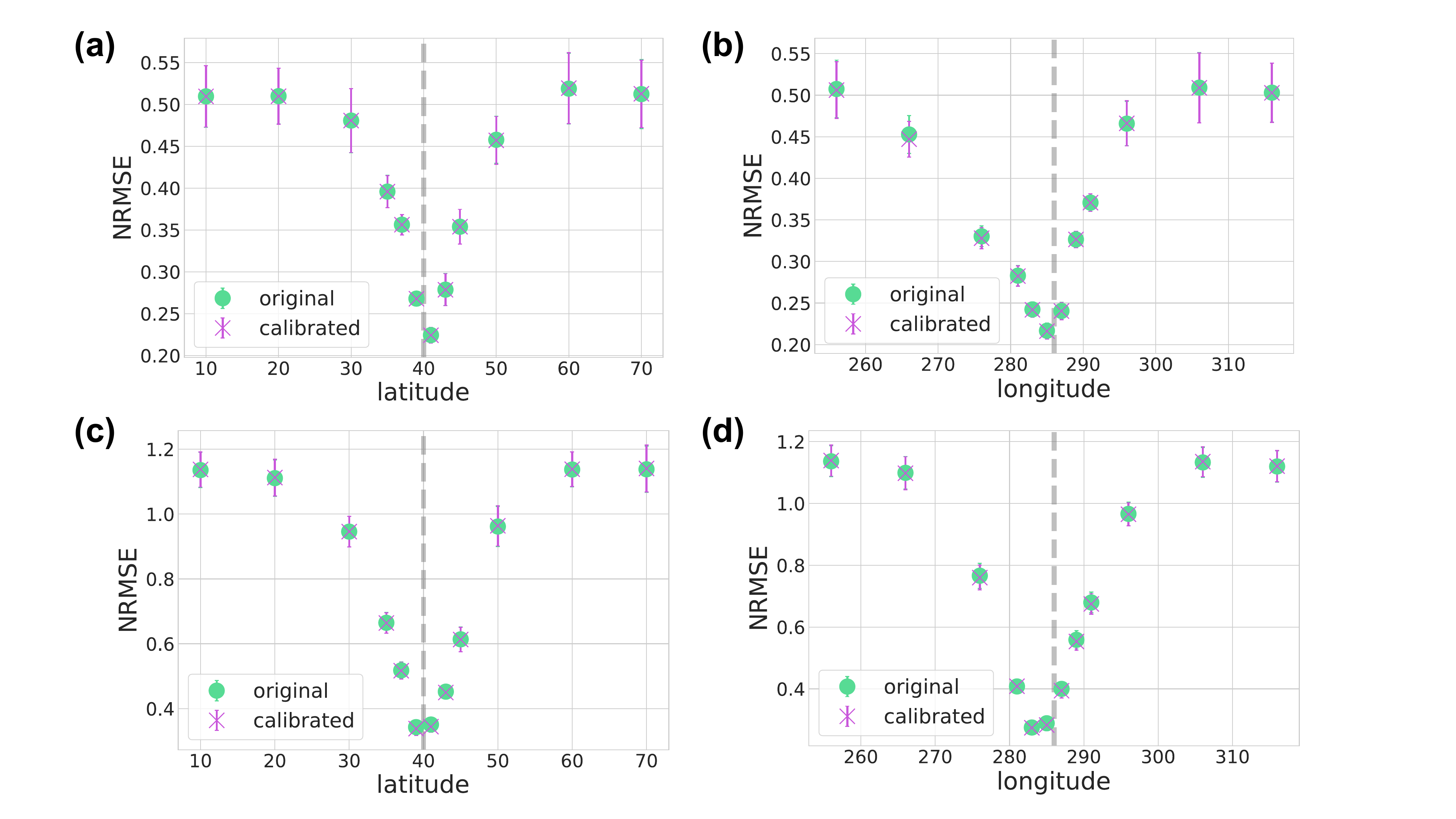}
  \vspace{-5pt}
 \caption{Results for the case where the target city is New York. The results of RC predictions using two different parameter sets are shown. Circles indicate the same parameter set as those in the main text (common to Tokyo) and the crosses indicate the calibrated parameter set.}\label{fig:calib_newyork}
\end{figure*}

\begin{figure*}[htb]
  \centering
  \includegraphics[keepaspectratio, width=1.8\columnwidth]
  {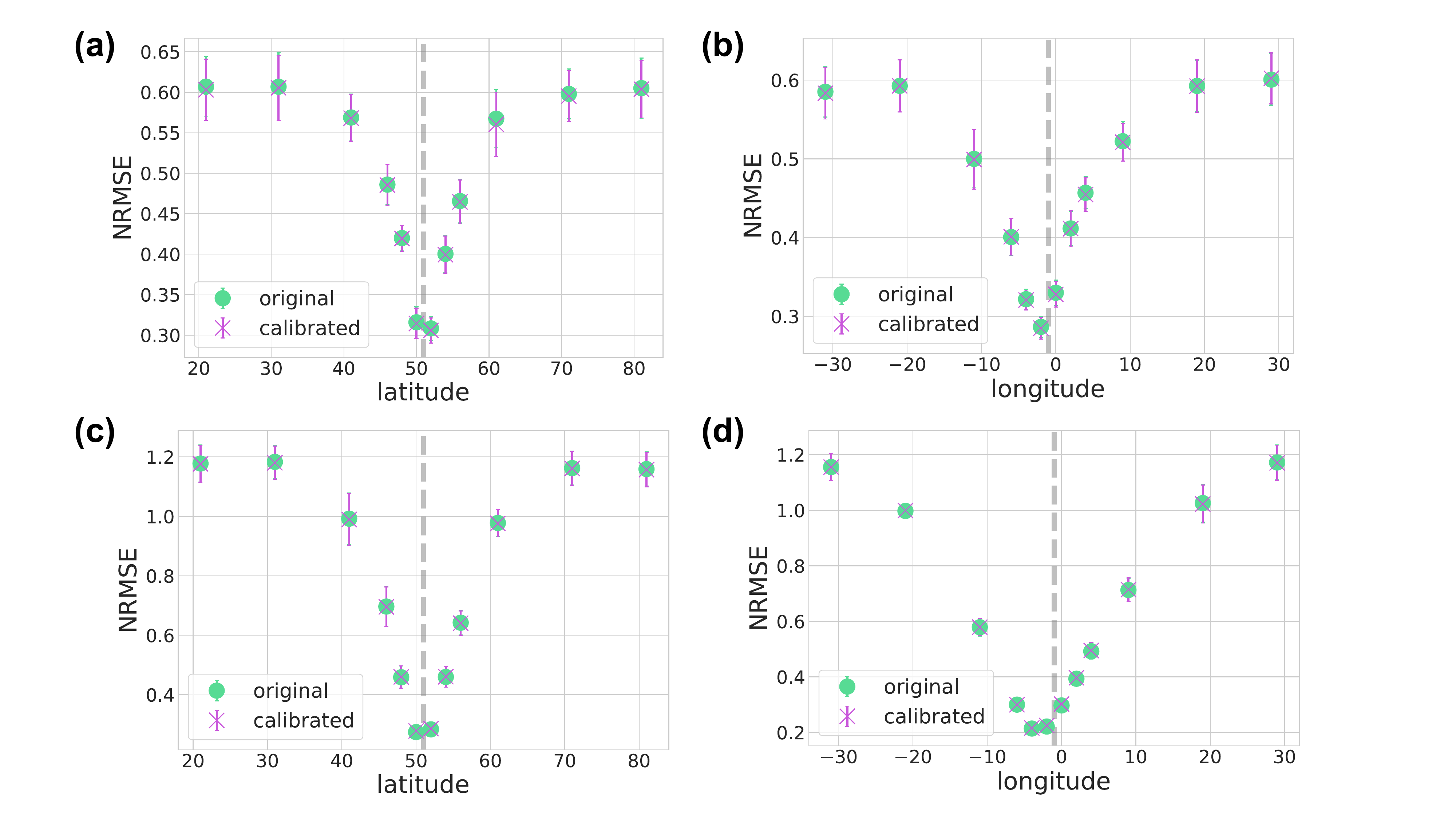}
  \vspace{-5pt}
 \caption{Results for the case where the target city is London. The quantities shown in this figure are the same as those in Fig.~\ref{fig:calib_newyork}.}\label{fig:calib_london}
\end{figure*}

\begin{figure*}[htb]
  \centering
  \includegraphics[keepaspectratio, width=1.8\columnwidth]
  {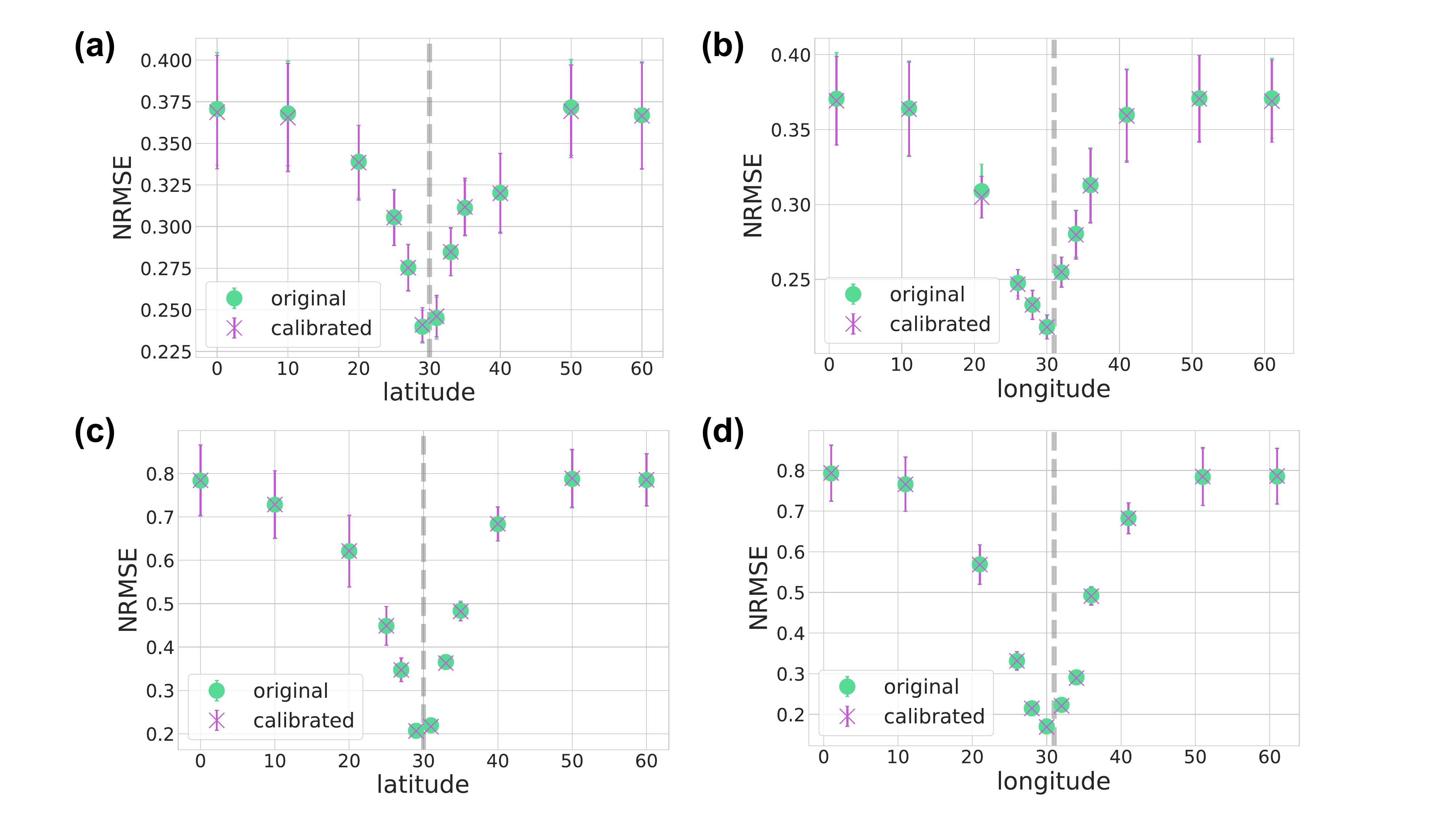}
  \vspace{-5pt}
 \caption{Results for the case where the target city is Cairo. The quantities shown in this figure are the same as those in Fig.~\ref{fig:calib_newyork}.}\label{fig:calib_cairo}
\end{figure*}

\begin{figure*}[htb]
  \centering
  \includegraphics[keepaspectratio, width=1.8\columnwidth]
  {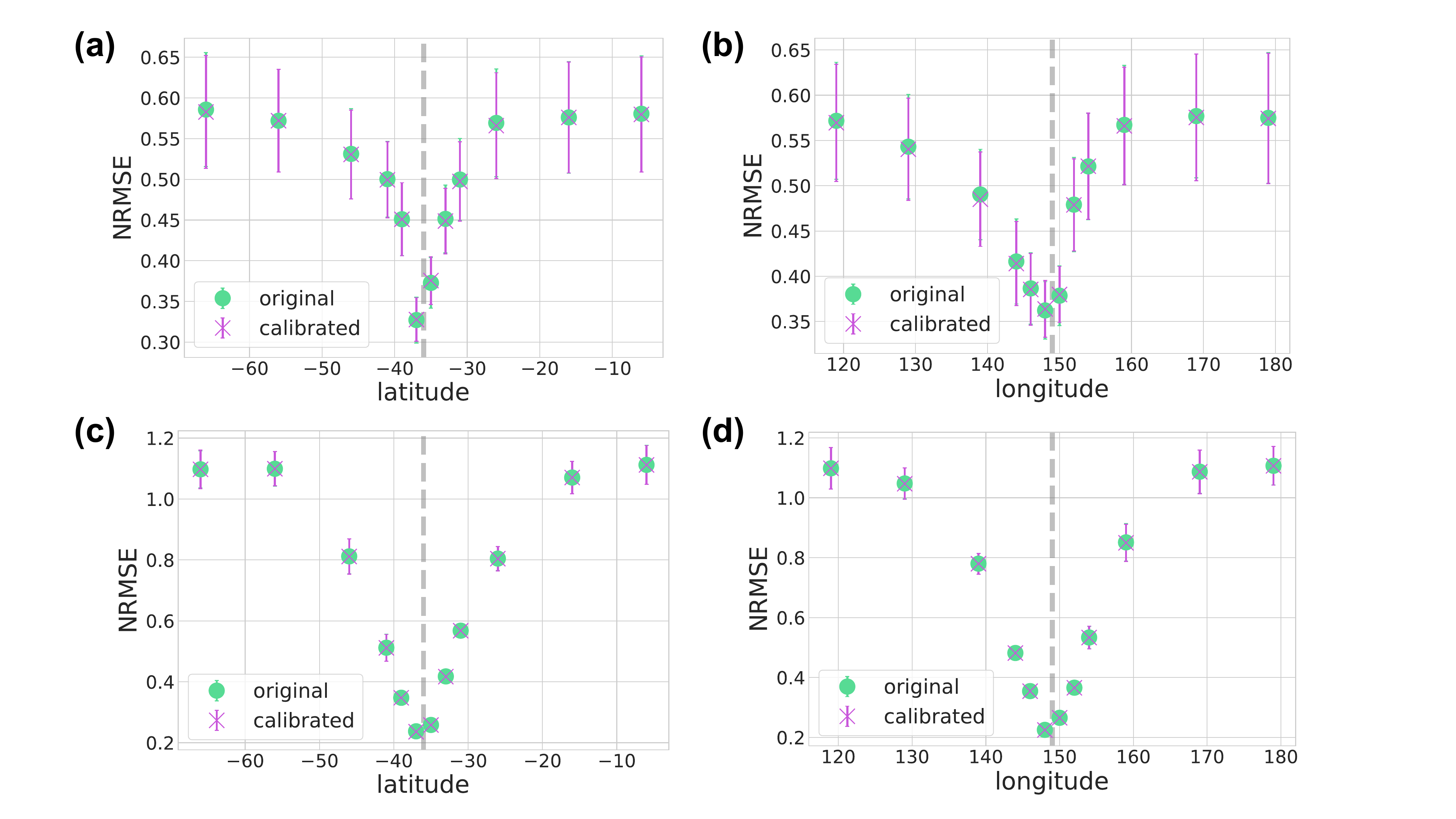}
  \vspace{-5pt}
 \caption{%予測地点がcanberraのケース．図の見方はFig.~\ref{fig:calib_newyork}と同様．
 Results for the case where the target city is Canberra. The quantities shown in this figure are the same as those in Fig.~\ref{fig:calib_newyork}.}\label{fig:calib_canberra}
\end{figure*}

We performed predictions by using RC with adjusted hyperparameters as listed in Table~\ref{table:param_calib}, for the cases where the target points are four cities; New York, London, Cairo, and Canberra. The parameter values that minimize the sum of the NRMSE for the points 1, 10, and 30$^\circ$ away from the target point were adopted as the optimal parameters. The conditions other than these four hyperparameters are the same as those in the main text. The prediction results using the hyperparameter values in the main text (i.e., those common to Tokyo) and the calibrated hyperparameter values are shown in Figs.~\ref{fig:calib_newyork}-\ref{fig:calib_canberra}. As can be seen from these figures, there was almost no effect of re-calibration in both cases, and it can be assumed that once the parameters are calibrated, they can be used to achieve good predictions for other points as well, at least for temperature and pressure prediction. In many cases, the values were partially similar, although in the data used here, there were no cases in which exactly the same set of parameters was used as a result of calibration. From these results, it may be possible to discuss in the future the approximate values of the parameters that can elicit good performance and their impact on prediction.
\begin{table}[htb]
\caption{The values of calibrated hyper parameters. Upper: for temperature data, lower: for pressure data.}\label{table:param_calib}
\centering
\begin{tabular}%{lcccc}\hline
{p{5em}p{5em}p{5em}p{5em}p{5em}}\hline
city & \hfil $d$\hfil & \hfil $\gamma$\hfil & \hfil $\rho$\hfil & \hfil $\beta$\hfil \\ \hline\hline
Tokyo & \hfil0.02\hfil & \hfil0.2\hfil & \hfil0.5\hfil & \hfil1.0\hfil \\
& \hfil0.07\hfil & \hfil0.05\hfil & \hfil0.2\hfil & \hfil0.15\hfil \\
New York & \hfil0.02\hfil & \hfil0.15\hfil & \hfil0.5 & \hfil0.45\hfil \\
& \hfil0.07\hfil & \hfil0.05\hfil & \hfil0.2\hfil & \hfil0.05\hfil \\
London & \hfil0.01\hfil & \hfil0.05\hfil & \hfil0.7 & \hfil1.0\hfil \\
& \hfil0.02\hfil & \hfil0.05\hfil & \hfil0.2\hfil & \hfil0.5\hfil \\
Cairo & \hfil0.02\hfil & \hfil0.05\hfil & \hfil0.5\hfil & \hfil0.85\hfil \\
& \hfil0.02\hfil & \hfil0.05\hfil & \hfil0.2\hfil & \hfil0.05\hfil \\
Canberra & \hfil0.02\hfil & \hfil0.05\hfil & \hfil0.6\hfil & \hfil1.0\hfil \\
& \hfil0.02\hfil & \hfil0.05\hfil & \hfil0.2\hfil & \hfil0.1\hfil \\ \hline
\end{tabular}
\end{table}
\section{Comparison between LI-ESN and ESN\label{sec:li-esn}}
\begin{figure*}[htb]
  \centering
  \includegraphics[keepaspectratio, width=1.9\columnwidth]
  {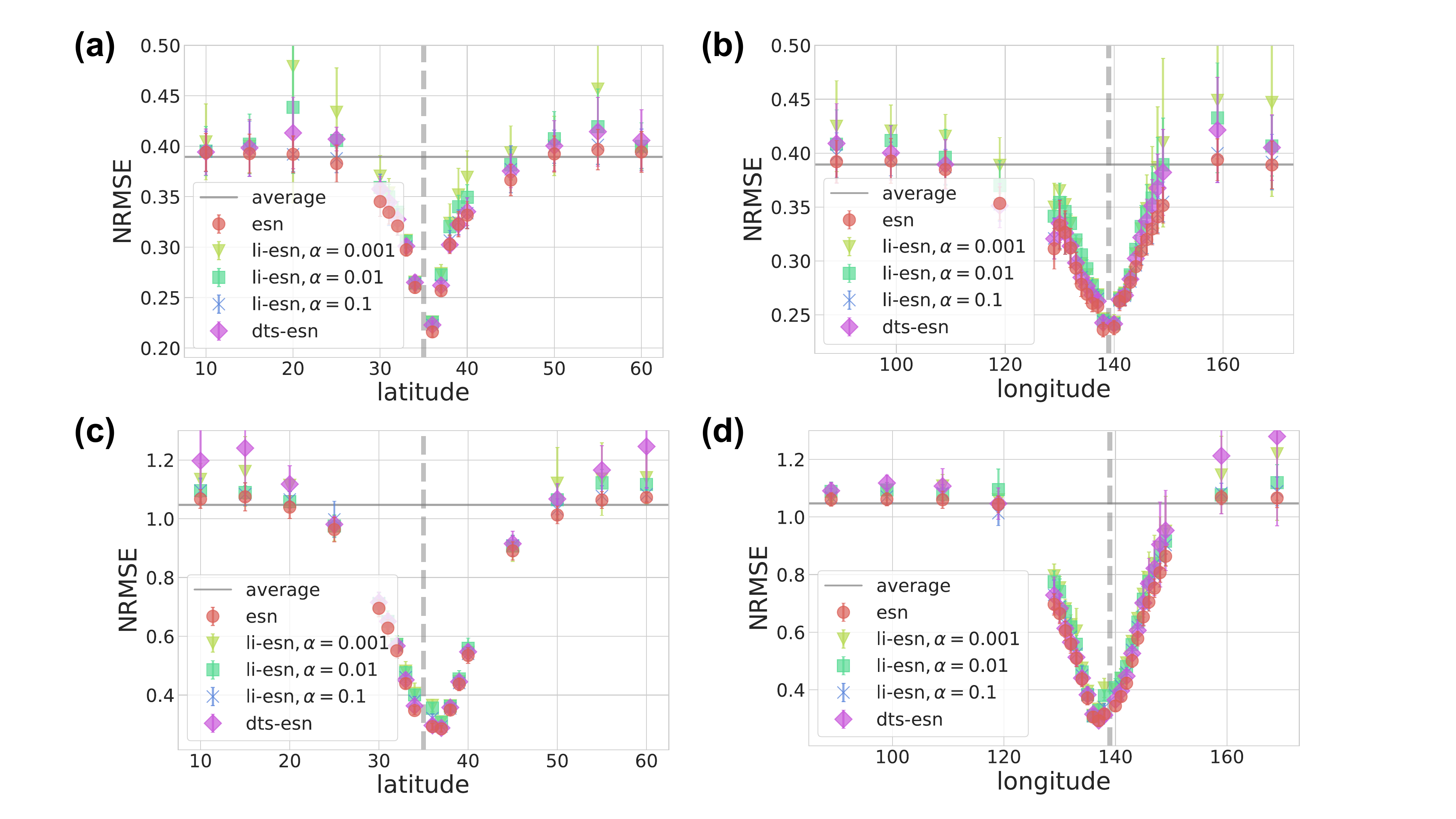}
  \vspace{-5pt}
 \caption{Results for the case where the target city is Tokyo. Results for LI-ESN with $\alpha=10^{-3},10^{-2},$ and $10^{-1}$, standard ESN, and DTS-ESN (Tokyo only) are shown. Predictors include ESN (circles), LI-ESN with $\alpha=10^{-3}$ (downward triangles), $10^{-2}$ (squares), $10^{-1}$ (crosses), and DTS-ESN (diamonds). The NRMSEs of historical average based prediction are plotted as gray horizontal line. (a) temperature, latitudinal. (b) temperature, longitudinal. (c) pressure, latitudinal. (d) pressure, longitudinal. The vertical gray dashed lines indicate the location of the target point.}\label{fig:liesn_tokyo}
\end{figure*}

\begin{figure*}[htb]
  \centering
  \includegraphics[keepaspectratio, width=1.9\columnwidth]
  {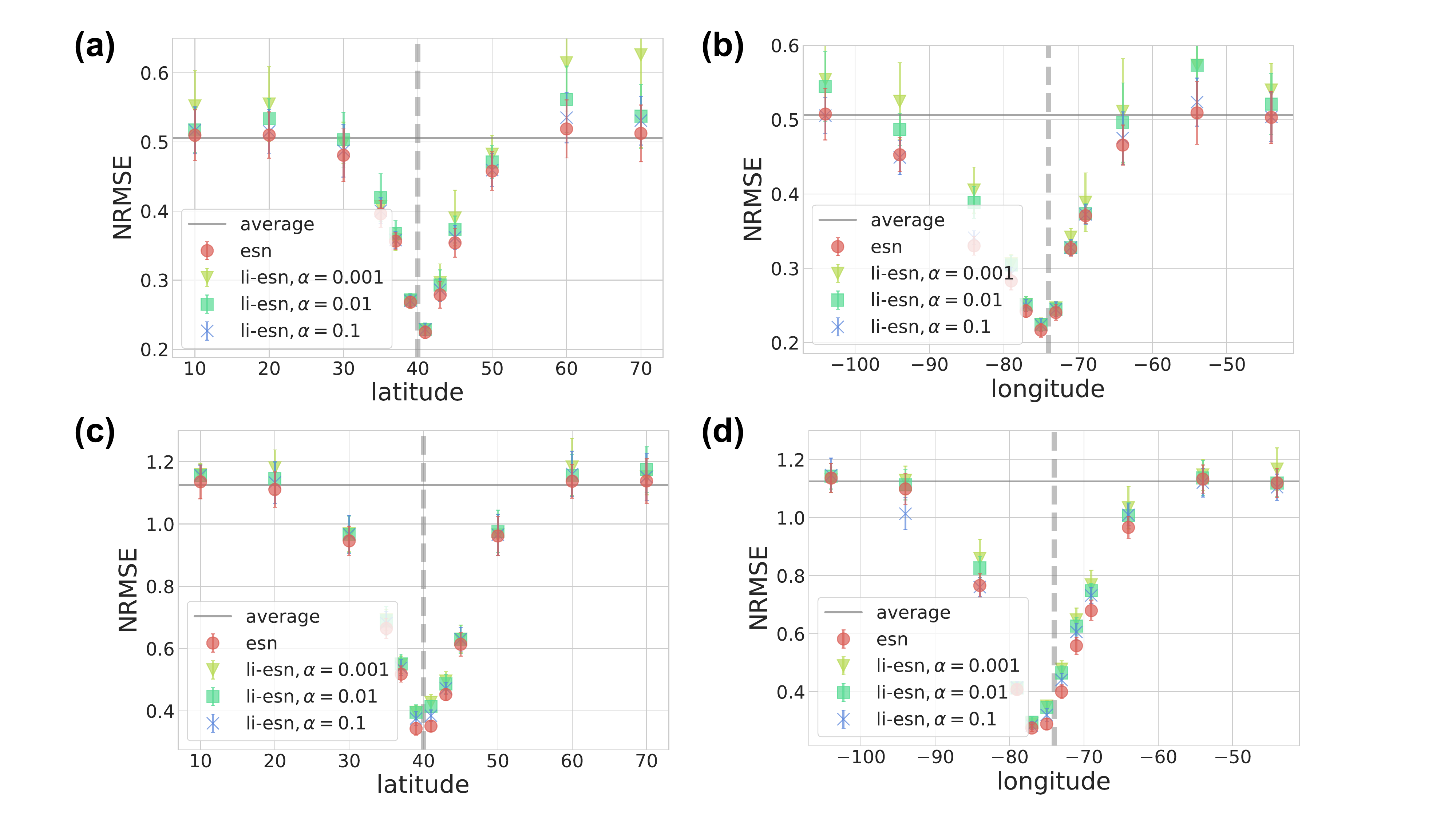}
  \vspace{-5pt}
 \caption{Results for the case where the target city is New York. The quantities shown in this figure are the same as those in Fig.~\ref{fig:liesn_tokyo}.}\label{fig:liesn_newyork}
\end{figure*}

\begin{figure*}[htb]
  \centering
  \includegraphics[keepaspectratio, width=1.9\columnwidth]
  {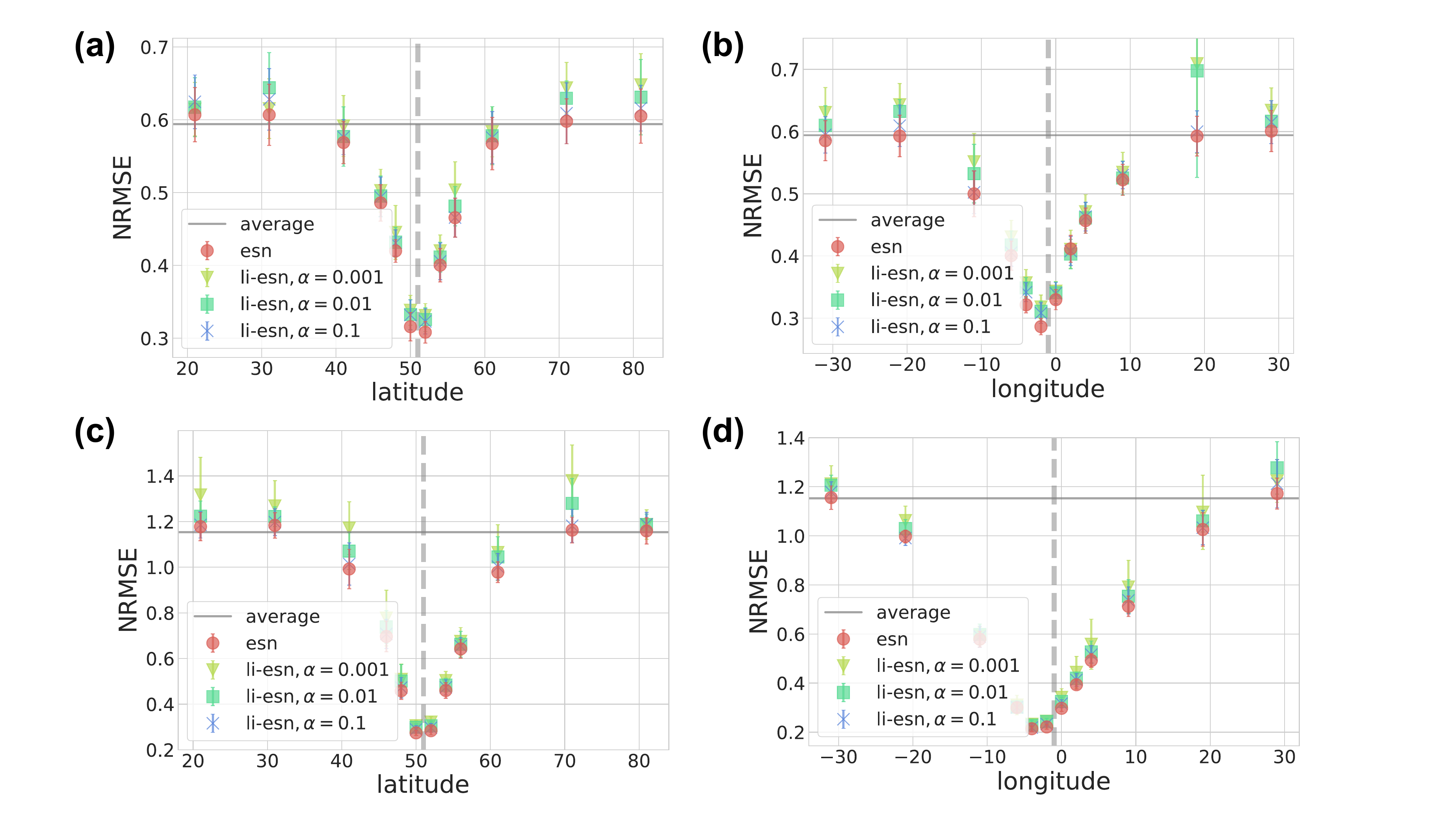}
  \vspace{-5pt}
 \caption{Results for the case where the target city is London. The quantities shown in this figure are the same as those in Fig.~\ref{fig:liesn_tokyo}.}\label{fig:liesn_london}
\end{figure*}

\begin{figure*}[htb]
  \centering
  \includegraphics[keepaspectratio, width=1.9\columnwidth]
  {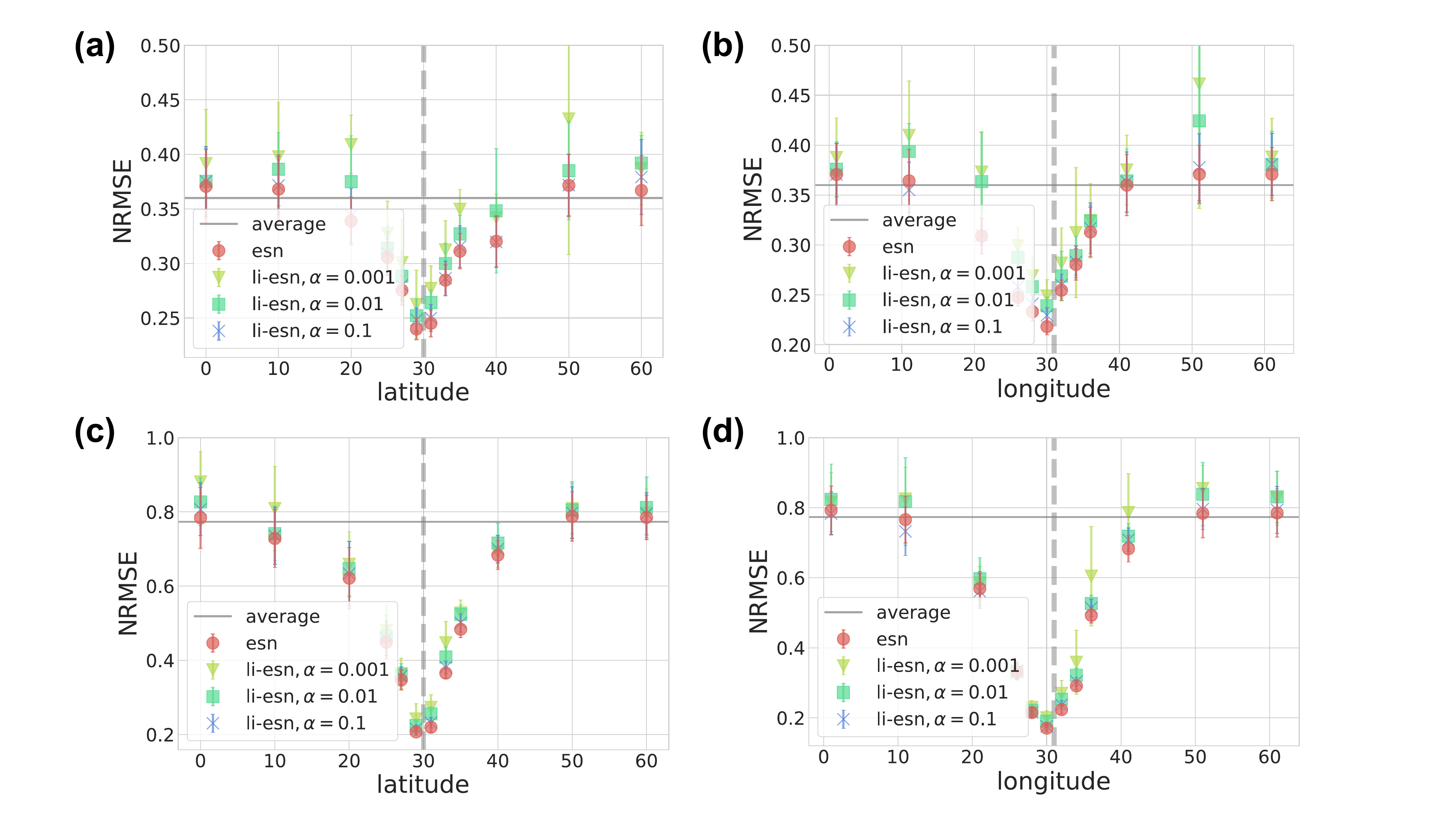}
  \vspace{-5pt}
 \caption{Results for the case where the target city is Cairo. The quantities shown in this figure are the same as those in Fig.~\ref{fig:liesn_tokyo}.}\label{fig:liesn_cairo}
\end{figure*}

In order to adapt to the time scales of various time series data variations, there is an improved RC called LI-ESN (Leakly Integrator Echo State Network)~\cite{JLP+2007} that can control the speed of model state fluctuations. The LI-ESN uses the Leakly Integrator (LI) model as a reservoir node, and the amount of state change is controlled by the leak rate $\alpha \in(0,1]$. When $\alpha=1$, the LI-ESN becomes the standard ESN.
In LI-ESN, the time evolution of the reservoir state vector ${\bf{x}}(n)$ and the output vector ${\bf{y}}(n)$ for the input vector ${\bf{u}}(n)$ at time $n$ are written as follows:
\begin{align}
\label{eq:esn1}
&{\bf{x}}(n) = (1-\alpha){\bf{x}}(n-1) \nonumber \\
&\;\;\;\;\;\;\;\;\;\;\;\;+ \alpha f(W^\mathrm{in}{\bf{u}}(n) + W{\bf{x}}(n-1)), \\
\label{eq:esn2}
&{\bf{y}}(n) = W^\mathrm{out}{\bf{x}}(n),
\end{align}
where $N_u,N_x (\gg N_u),N_y$ are the number of nodes in the input layer, reservoir and output layers, and $W^\mathrm{in},W,W^\mathrm{out}$ are the joint weight matrix between the input and reservoir layers, within the reservoir, and between the reservoir and output layers, respectively. Here, as with the ESN in the main text, the activation function is $f=\tanh$, and the ridge regression is employed as learning algorithm. The hyperparameters other than $\alpha$ are common to ESN.
Time-series data were predicted using LI-ESN with $\alpha=10^{-3},10^{-2},$ and $10^{-1}$. The dataset used were the same as in the main text, with five target cities. As shown in Fig.~\ref{fig:liesn_tokyo}-\ref{fig:liesn_cairo}, LI-ESN exhibited larger NRMSE than ESN, and the NRMSE became larger as $\alpha$ became smaller. The smaller $\alpha$ corresponds to a slower change in the reservoir's state vector, indicating that a predictor that can handle fast time changes is appropriate for our prediction.

Furthermore, there is an extended model of LI-ESN, DTS-ESN~\cite{TMY+2022}. This model can handle time-series data that include data with different time scales of variation by having a distribution for $\alpha$, which is common to all nodes in the LI-ESN. We used DTS-ESN to predict time series data targeting Tokyo. The leak rate was set to be $\log_{10}\alpha_i \in[-3,0]$. Here, the hyperparameters were calibrated values of Table~\ref{table:param_dtsesn}.
Fig.~\ref{fig:liesn_tokyo} shows that the standard ESN still performs better than the DTS-ESN. Based on the LI-ESN discussion, a predictor with a large $\alpha$, i.e., one that can capture fast time scales, is preferred for this data set, and the DTS-ESN results are consistent with this discussion. The distribution of $\alpha$ in DTS-ESN includes the range of $\alpha$ set in ESN and LI-ESN. This confirms that it is difficult to improve the prediction performance by setting the $\alpha$ value of a node individually if it is not a $\alpha$ of good nature.
\begin{table}[htb]
\caption{The values of calibrated hyperparameter values for DTS-ESN (Upper: temperature data, lower: pressure data).}\label{table:param_dtsesn}
\centering
\begin{tabular}%{lcccc}\hline
{p{5em}p{5em}p{5em}p{5em}p{5em}}\hline
city %& hyper parameter (temperature,pressure)\\  
& \hfil $d$\hfil & \hfil $\gamma$\hfil & \hfil $\rho$\hfil & \hfil $\beta$\hfil \\ \hline\hline
Tokyo & \hfil0.01\hfil & \hfil0.05\hfil & \hfil0.1\hfil & \hfil1.0\hfil \\
& \hfil0.02\hfil & \hfil0.05\hfil & \hfil0.3\hfil & \hfil1.0\hfil \\ \hline
\end{tabular}
\end{table}
%
% \section{The reration between NRMSE and input-output correlation}
% The relationship between NRMSE and input-output data correlation was examined for all the target cities examined in this study. Scatter plots for (a) temperature and (b) pressure are shown in Fig.~\ref{fig:corr_vs_nrmse}. The results for all the target cities show a positive correlation, even though the values differ depending on the city. Furthermore, the figure shows that the VAR tends to have a larger error than the RC in cases where the correlation is high (when the correlation coefficient $\gtrsim$ 0.6). This reflects the fact that RC is more stable when the input-output data are similar, as mentioned in the main text.
%
\newpage
\begin{figure*}[htb]
  \centering
  \includegraphics[keepaspectratio, width=1.9\columnwidth]
  {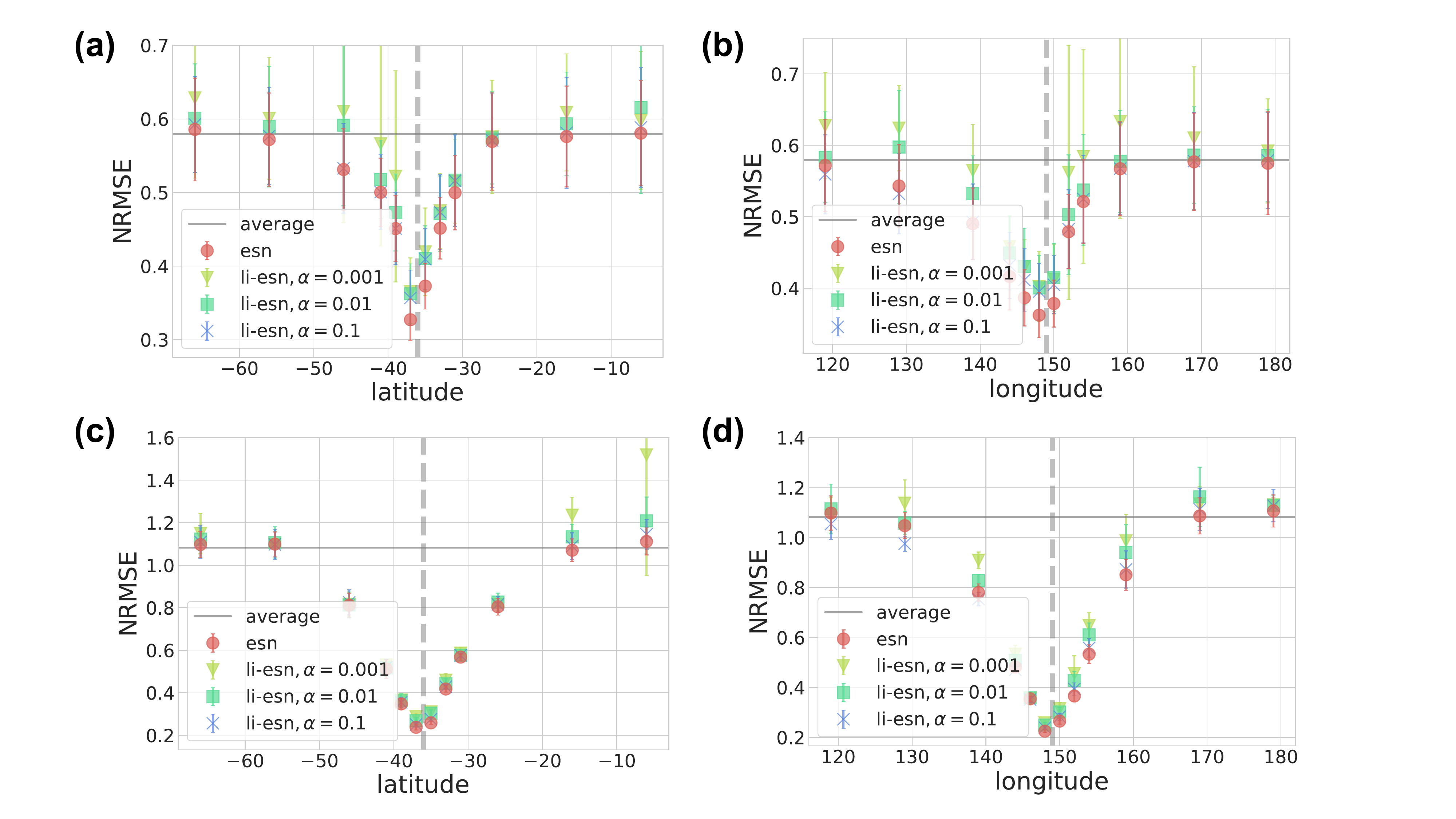}
  \vspace{-5pt}
 \caption{Results for the case where the target city is Canberra. The quantities shown in this figure are the same as those in Fig.~\ref{fig:liesn_tokyo}.}\label{fig:liesn_canberra}
\end{figure*}
\clearpage
\bibliography{arXiv_main}
\end{document}